\pdfoutput=1

\documentclass[11pt]{article}

\usepackage{emnlp2021}

\usepackage{times}
\usepackage{latexsym}

\usepackage[T1]{fontenc}

\usepackage[utf8]{inputenc}

\usepackage{microtype}

\usepackage{booktabs}
\usepackage{todonotes}
\usepackage{amsmath}
\usepackage{bbm}
\usepackage{amssymb}
\usepackage{multirow}
\usepackage{makecell}
\usepackage{subcaption}
\usepackage{enumitem}

\title{A Comprehensive Assessment of Dialog Evaluation Metrics}

\author{Yi-Ting Yeh, Maxine Eskenazi, Shikib Mehri \\
  Language Technologies Institute, Carnegie Mellon University \\
  \texttt{\{yitingye,max,amehri\}@cs.cmu.edu}}

\begin{document}
\maketitle
\begin{abstract}
Automatic evaluation metrics are a crucial component of dialog systems research. Standard language evaluation metrics are known to be ineffective for evaluating dialog. As such, recent research has proposed a number of novel, dialog-specific metrics that correlate better with human judgements. Due to the fast pace of research, many of these metrics have been assessed on different datasets and there has as yet been no time for a systematic comparison between them. To this end, this paper provides a comprehensive assessment of recently proposed dialog evaluation metrics on a number of datasets. In this paper, 23 different automatic evaluation metrics are evaluated on 10 different datasets. Furthermore, the metrics are assessed in different settings, to better qualify their respective strengths and weaknesses. Metrics are assessed (1) on both the turn level and the dialog level, (2) for different dialog lengths, (3) for different dialog qualities (e.g., coherence, engaging), (4) for different types of response generation models (i.e., generative, retrieval, simple models and state-of-the-art models), (5) taking into account the similarity of different metrics and (6) exploring combinations of different metrics. This comprehensive assessment offers several takeaways pertaining to dialog evaluation metrics in general. It also suggests how to best assess evaluation metrics and indicates promising directions for future work.

\end{abstract}

\section{Introduction}

Evaluation is a crucial component of the research process. Evaluation metrics are used to shine light on the best models and thus they strongly influence the research directions of a field. Standard automatic language evaluation metrics (e.g., BLEU, METEOR) have been shown to be ineffective for evaluating dialog \citep{liu2016not,deriu2021survey}. To this end, recent research has proposed a number of automatic metrics specifically designed to evaluate dialog \citep{tao2018ruber,ghazarian-etal-2019-better,10.1145/3423168,pang-etal-2020-towards,ghazarian2020predictive,sinha-etal-2020-learning,huang-etal-2020-grade,mehri-eskenazi-2020-usr,mehri-eskenazi-2020-unsupervised,zhang2021deep}. These metrics address the weaknesses of the standard language evaluation metrics. They have also been shown to better correlate with human judgement. However, most of these metrics were developed at about the same time and were evaluated on different datasets. As such, there has not yet been a consistent comparison amongst them. This paper describes an assessment of these metrics on several different datasets. It quantifies the relative performance and reveals strengths and weaknesses of each metric. 

Due to the fact that standard automatic metrics have been shown to be ineffective for dialog evaluation \citep{liu2016not,deriu2021survey}, dialog research typically relies on human evaluation. While useful and very informative, human evaluation is expensive and time-consuming. Given these considerations, at present it is usually only used for a final evaluation. Yet, during the development process, automatic evaluation metrics are essential since they are used to optimize the model design and the choice of hyperparameters. In order to be relevant and meaningful, these metrics must better correlate with human judgment so that they serve as a meaningful proxy for human judgements during the development process.

Developing meaningful automatic evaluation metrics for dialog is challenging for several reasons. 
\begin{enumerate}
    \item The one-to-many nature of dialog \citep{zhao2017learning} makes word-overlap metrics ineffective. There are many appropriate responses to a particular dialog context and a system output should not be penalized for deviating from the ground-truth response. To mitigate this problem, several recently-proposed automatic metrics are \textit{reference-free}. Instead of comparing the generated response to the ground-truth response, these metrics evaluate the generated response in the context of the dialog history. 
    \item Since the set of conversation topics in open-domain dialog is unbounded, dialog evaluation metrics must be able to model the semantic meaning of the dialog context and the generated response for a variety of different topics. To address this problem, the majority of recently proposed metrics rely on pretrained language models \citep{devlin-etal-2019-bert} and self-supervised training objectives. 
    \item  The use of labeled data to train dialog evaluation metrics has several problems. First, collecting labeled data is expensive and is therefore almost always only done for the assessment of evaluation metrics. More importantly, training on labeled data could drastically limit the scope of an evaluation metric. A metric that is trained on quality-annotated data may over-optimize on the training data, both in terms of the conversation topics and the response generation models in the dataset. As such, many recently-proposed metrics design self-supervised training objectives that aim to capture different properties of dialog (e.g., relevance, fluency, etc.). Though these metrics are generally trained on specific datasets, the use of self-supervised objectives allows for future adaptation to new topics.
\end{enumerate}

This paper discusses the assessment of several recently-proposed automatic evaluation metrics for dialog over several datasets. These datasets have human annotations that measure the quality of the responses. Several quality annotated datasets are used: USR \citep{mehri-eskenazi-2020-usr}, GRADE \citep{huang-etal-2020-grade}, HolisticEval \citep{pang-etal-2020-towards}, FED \citep{mehri-eskenazi-2020-unsupervised}, DSTC6 \citep{DSTC6_End-to-End_Conversation_Modeling} and DSTC9 \citep{gunasekara2020overview}. Furthermore, this paper includes analysis experiments that qualify the performance of the metrics across different settings. This assessment measures the performance of the metrics (1) on both the turn level and the dialog level, (2) for different dialog lengths, (3) for different dialog qualities (e.g., coherence, engaging), (4) for different types of response generation models (i.e., generative, retrieval, simple models and state-of-the-art models), (5) taking into account the similarity of different metrics and (6) exploring combinations of different metrics. 

All of the code created for this assessment is open-sourced\footnote{\url{https://github.com/exe1023/DialEvalMetrics}} to facilitate the easy assessment of future evaluation metrics on a number of datasets.

\section{Task Definition}
A variety of metrics are assessed here that automatically evaluate responses produced by dialog systems.
Formally, given a dialog context $c=\{c_1, \dots, c_m\}$, model response $r=\{r_{1}, \dots, r_n\}$, and human written reference response $g=\{g_{1}, \dots, g_{k}\}$, the goal is to learn a function $f: (c, r, g) \rightarrow s$ that evaluates the generated response. A metric is said to be reference-free if it can be expressed as a function of \textit{only} $c$ and $r$.

These metrics are assessed by comparing them to human judgment. Concretely, a human annotator (or several annotators) scored the quality of a given response conditioned on the dialog context: $(c,r) \rightarrow q$. Given the scores produced by a particular metric, $S = \{ s_1, s_2, \dots, s_n \}$, and the corresponding human quality annotations, $Q = \{ q_1, q_2, \dots, q_n \}$, we can measure the performance of the metric by calculating the correlation between $S$ and $Q$.

\begin{table*}[]
\small
    \centering
    \begin{tabular}{l|cccc}
    \toprule
    Metric & Pretrained Model & Training Dataset  & Reference-Free? & Objective \\
    \midrule
    BLEU \shortcite{papineni-etal-2002-bleu} & X &  X & X & X \\
    METEOR \shortcite{banerjee-lavie-2005-meteor} & X &  X & X & X \\
    ROUGE \shortcite{lin-2004-rouge} & X &  X & X & X \\
    ADEM \shortcite{lowe-etal-2017-towards} & X & Ubuntu Dialogue + Twitter & X & MSE \\
    BERTScore \shortcite{zhang2019bertscore} & BERT & X & X & X \\
    BLEURT \shortcite{sellam-etal-2020-bleurt} & BERT & WMT Metrics Shared Task & X & MSE \\
    QuestEval \shortcite{scialom2021questeval} & T5 &  SQuAD-v2/NewsQA & $\surd$  & QA/QG \\
    RUBER \shortcite{tao2018ruber} & X & DailyDialog / PersonaChat & X & Triplet  \\
    BERT-RUBER \shortcite{ghazarian-etal-2019-better} & BERT & DailyDialog / PersonaChat & X & Triplet \\
    PONE \shortcite{10.1145/3423168} & BERT & DailyDialog & X & Triplet \\
    MAUDE \shortcite{sinha-etal-2020-learning} & BERT & PersonaChat & $\surd$ & NCE \\
        DEB  \shortcite{sai-etal-2020-improving} & BERT & Reddit/DailyDialog++ & $\surd$ & MLM/NSP \\
        GRADE \shortcite{huang-etal-2020-grade} & BERT & DailyDialog & $\surd$ & Triplet  \\
        DynaEval \shortcite{zhang-etal-2021-DynaEval} & RoBERTa & ED/ConvAI2/DailyDialog & $\surd$ & Triplet \\
        USR \shortcite{mehri-eskenazi-2020-usr} & RoBERTa & TopicalChat / PersonaChat & $\surd$ & MLM/CrossEntropy \\
        USL-H \shortcite{phy-etal-2020-deconstruct} & BERT & DailyDialog & $\surd$ & VUP/NSP/MLM \\
        DialogRPT    \shortcite{gao2020dialogrpt} & GPT-2 & Reddit & $\surd$ & CrossEntropy \\
        Deep AM-FM \shortcite{zhang2021deep} & Multilingual BERT & Twitter & X & MLM \\ 
    HolisticEval \shortcite{pang-etal-2020-towards} & BERT & DailyDialog & $\surd$ & LM \\
    PredictiveEngage \shortcite{ghazarian2020predictive} & BERT & ConvAI & X & CrossEntropy \\
    FED \shortcite{mehri-eskenazi-2020-unsupervised} & DialoGPT & X & $\surd$ & X \\
    FlowScore \shortcite{li2021dialoflow} & Plato & Reddit & $\surd$ & ContextFlow \\
    FBD \shortcite{xiang2021assessing} & RoBERTa & X & X & X \\

     \end{tabular}
    \caption{Summary of the evaluation metrics assessed in this paper. The `Pretrained Model' column indicates the specific pretrained language model used by the metric. The `Training Dataset' and `Objective' columns describe the dialog data and the objective used when training the metric. `Reference-Free?' indicates whether the metric requires a reference response for evaluation. ED is the abbreviation of the EmpatheticDialogue dataset. }
    \label{tab:metric_overview}
\end{table*}

\begin{table*}[]
    \tiny
    \centering
    \begin{tabular}{ll}
    \toprule
    Metrics & Reason for not using \\
    \midrule
UNION \shortcite{guan-huang-2020-union} & Designed for story generation \\
Embedding Methods \shortcite{kiros2015skip, liu-etal-2016-evaluate, rodriguez2021automatic} & Computing the embedding similarity is identical to the referenced metric in RUBER. \\
Language Model Evaluator \shortcite{nedelchev-etal-2020-language} & Using the language model probability is identical to the approach of USR and HolisticEval. \\
Perplexity \shortcite{Adiwardana2020TowardsAH} & No access to the NLG models that generated the quality-annotated data in order to obtain perplexity. \\
Distinct N-grams \shortcite{li-etal-2016-diversity} & This is not feasible for models trained on dialog corpora \cite{pang-etal-2020-towards}. \\
Spot The Bot \shortcite{deriu-etal-2020-spot} & This still needs human annotation when evaluating models. \\
Learning to Compare \shortcite{zhou2020learning} & No released code, data, or pretrained models. \\
RUSE \shortcite{shimanaka-etal-2018-ruse} & Designed for machine translation only. \\
uBLEU \shortcite{yuma-etal-2020-ubleu} & No released code. \\
deltaBLEU \shortcite{galley-etal-2015-deltableu} & For each response, this requires multiple human references to calculate the score. \\
Data-QuestEval \shortcite{Rebuffel2021DataQuestEvalAR} & Designed for general NLU. not specifically for dialog.  \\
Topic-based Evaluation \shortcite{Guo2018TopicbasedEF} & No released code, data, or pretrained models. \\
The Alexa Prize Evaluation Framework \shortcite{venkatesh2018evaluating} & No released code, data, or pretrained models. \\
AIH \shortcite{li2021addressing} & No released code, data, or pretrained models.
    \end{tabular}
\caption{The metrics that were not assessed in this paper and the reasons behind that choice.}
\label{tab:not_used_metrics}
\end{table*}

\section{Overview of Automatic Metrics}

This section describes the automatic metrics assessed in this paper. Table~\ref{tab:metric_overview} presents an overview of these metrics. The three aspects used to characterize them are: (1) Does the metric use a pretrained language model? (2) What data was used to train the metric? (3) Does the metric require a reference response or is it reference-free? 

\subsection{Design of the Metrics in this Assessment}

In general, dialog evaluation metrics can be divided into rule-based and model-based metrics. Rule-based metrics use heuristic rules to evaluate the system response, conditioned on the dialog context and human reference(s). Model-based metrics are trained, often with self-supervised objectives, to measure the quality of the responses.

BLEU~\cite{papineni-etal-2002-bleu} is a popular rule-based metric often used to benchmark natural language generation (NLG) systems. BLEU computes the n-gram precision of the system responses using human references. 
While BLEU performs reasonably when evaluating NLG systems, it has issues reflecting grammaticality, use of semantically similar words, and meaning preservation \cite{novikova-etal-2017-need, ananthakrishnan2007some, sulem-etal-2018-bleu, reiter-2018-structured}.

METEOR~\cite{banerjee-lavie-2005-meteor} and ROUGE~\cite{lin-2004-rouge} have been proposed to address the shortcomings of BLEU.
METEOR incorporates stems and synonyms into its calculation, while ROUGE focuses on n-gram recall instead of precision.
Although these two metrics improve upon BLEU, they remain ineffective for dialog evaluation \citep{liu2016not}. In general, word-overlap metrics struggle to evaluate dialog responses because of the one-to-many nature of dialog \citep{zhao2017learning}. Concretely, as mentioned above, since there are many appropriate responses for a given dialog context it is unreasonable to penalize a valid system response that deviates from the ground truth.

ADEM~\cite{lowe-etal-2017-towards} is an early learning-based metric that uses a recurrent neural network (RNN) to predict the quality of system responses. ADEM uses quality-annotated training data and the model is trained to predict human quality annotations with a mean squared error loss (MSE).

RUBER~\cite{tao2018ruber} uses a hybrid model consisting of both a referenced metric and an unreferenced metric.
The referenced metric calculates the cosine similarity of word embeddings between a system response and a human reference. The unreferenced metric is trained with a triplet ranking loss to predict whether the generated response is appropriate for the dialog history. BERT-RUBER~\cite{ghazarian-etal-2019-better} proposes to replace the RNN in RUBER with BERT~\cite{devlin-etal-2019-bert} to further improve the performance with contextualized word embeddings.

Based on BERT-RUBER, PONE~\cite{10.1145/3423168} uses a novel algorithm to sample negative examples during training, and trains the metric on a dataset augmented by other NLG models.

MAUDE~\cite{sinha-etal-2020-learning} is trained with Noise Contrastive Estimation (NCE)~\cite{gutmann2010noise} which requires the model to differentiate between a correct response and randomly sampled negative responses. 

DEB~\cite{sai-etal-2020-improving} constructs a dialog dataset which consists of manually-created relevant responses and adversarial irrelevant responses. 
DEB uses BERT for dialog evaluation by first pretraining on a large-scale dialog corpus, and then fine-tuning on the proposed dataset with a next sentence prediction (NSP) objective.

GRADE~\cite{huang-etal-2020-grade} models topic transition dynamics in dialog by constructing a graph representation of the dialog history. This graph is then passed as input to a model that is trained with the same triplet loss as RUBER. 

While GRADE is focused on turn-level topic transition dynamics in dialog, DynaEval~\cite{zhang-etal-2021-DynaEval} leverages a graph structure to model the dialog-level interactions between a user and a system. Through this graph-based approach, DynaEval is trained to distinguish well-formed dialogs from carefully constructed negative samples.

USR~\cite{mehri-eskenazi-2020-usr} trains several models to measure different qualities of dialogs. USR relies on three different models, (1) a language model, trained with the masked language modelling (MLM) objectives, measures fluency, (2) a dialog retrieval model determines the relevance of a response and (3) a fact-to-response selection model measures whether a response conditions on knowledge.

Similarly, USL-H~\cite{phy-etal-2020-deconstruct} combines three models trained with different objectives: valid utterance prediction (VUP), next sentence prediction (NSP), and MLM. 
The VUP model determines whether a response is valid and grammatically correct. 
The NSP model and MLM models are trained with self-supervised objectives to evaluate the sensibleness and the likelihood of a given response. 

DialogRPT~\cite{gao2020dialogrpt} is an ensemble of multiple GPT-2 based models, which were fine-tuned on the Reddit human feedback data with different tasks. The tasks include predicting human feedback of responses and whether the response is human-like.

The Deep AM-FM metric~\cite{zhang2021deep} measures two aspects of dialog quality through the Adequacy Metric (AM) and the Fluency Metric (FM).
AM assesses the semantic similarity of system responses and human references by comparing their BERT embeddings.
FM compares the similarity of the language model probabilities for both the system response and the human reference, and produces a higher score if the probabilities are similar.

HolisticEval~\cite{pang-etal-2020-towards} evaluates several qualities of dialog: \textit{context coherence}, \textit{language fluency}, \textit{response diversity}, and \textit{logical self-consistency}. The GPT-2 language  model~\cite{radford2019language} and pretrained Natural Language Inference models are used to measure these qualities. 

In addition to measuring the relevance of a response, PredictiveEngage~\cite{ghazarian2020predictive} incorporates an utterance-level engagement classifier to better assess the overall quality of a response.

FED~\cite{mehri-eskenazi-2020-unsupervised} is an unsupervised evaluation metric that uses DialoGPT~\cite{zhang2019dialogpt} to measure 18 fine-grained qualities of dialog. FED calculates the likelihood of manually designed follow-up utterances to measure multiple qualities of dialog without any supervision.

FlowScore~\cite{li2021dialoflow} which is based on the DialoFlow model, models the dynamic information flow in the dialog history to evaluate the quality of a dialog. DialoFlow is a response generation model that is trained with three objectives CFM, SIM and RGM in order to condition the response generation on the context flow in the dialog. FlowScore uses the representations produced by DialoFlow to measure the dialog quality.

FBD~\cite{xiang2021assessing} computes the distribution-wise difference between the system generated conversations and the human-written conversations to evaluate the performance of a dialog system.
FBD focuses on assessing system-level performance and leverages the pretrained RoBERTa model without any additional fine-tuning.

In addition to automatic metrics specifically designed for dialog evaluation, this paper also evaluates the performance of BERTScore~\cite{zhang2019bertscore}, QuestEval~\cite{scialom2021questeval}, and BLEURT~\cite{sellam-etal-2020-bleurt} which were originally designed for evaluating machine translation, summarization and general natural language generation. 
BERTScore computes the F1 score by matching token embeddings in the human reference and system response. BLEURT generates synthetic data to pre-train BERT and fine-tune the model to predict a human score with MSE loss.  QuestEval, which is based on question generation (QG) and question answering (QA), accounts for factual consistency, relevance, and information selection of the generated response.
While these three metrics were not specifically designed for dialog, it is interesting to observe how they perform on dialog data.

\subsection{Use of Pretrained Language Models}

Large-scale pretrained language models~\cite{devlin-etal-2019-bert,radford2019language} have been ubiquitous in NLP models. Embeddings from pretrained language models have been shown to be particularly effective for a variety of NLP tasks. Pre-trained models are now a commonly-used strategy in dialog evaluation metrics.
However, since different pretrained models use different training data and objectives, the choice of language model might significantly influence the final performance and generalizability of the evaluation metrics. \textit{Future work should explore the impact of the different pretrained language models on the performance of the evaluation metric.}

BERT \citep{devlin-etal-2019-bert} is used in many of the metrics that this paper assesses. BERT uses the Masked Language Modeling (MLM) and Next Sentence Prediction (NSP) objectives, and is trained on the BookCorpus \cite{zhu2015aligning} and English Wikipedia.

RoBERTa, which is employed in USR~\cite{mehri-eskenazi-2020-usr}, improves the training techniques in BERT and trains the model on a much larger corpus which includes the CommonCrawl News dataset~\cite{mackenzie2020cc} and text extracted from Reddit. 
Specifically, the full training data size of RoBERTa is 10 times larger than BERT, and empirically RoBERTa has better performance than BERT on common NLP tasks including the GLUE benchmark~\cite{wang-etal-2018-glue}.

FED~\cite{mehri-eskenazi-2020-unsupervised} uses DialoGPT~\cite{zhang2019dialogpt} which is based on the GPT2~\cite{radford2019language} architecture and trained with dialog data extracted from Reddit.
Due to this, DialoGPT might better model human conversation, particularly open-domain chit-chat.

Deep AM-FM uses Multilingual BERT \cite{devlin-etal-2019-bert}, which is BERT trained on multilingual datasets. \textit{The benefits of using multilingual language models to evaluate English data is unclear and is an interesting topic for future work.}

T5~\cite{2020t5}, which is used by QuestEval, is a Transformer Seq2Seq model pretrained on a massive text corpus.
T5 achieved state-of-the-art results on many text generation tasks including summarization, question answering, and text classification.
The strong performance of T5 on NLG tasks, suggests that it may be valuable in the assessment of response generation models.

\subsection{Training Data} \label{sec:train_data}

The choice of training data is one of the most important factors in model-based metrics. The domain, quality, and conversation setting of the dataset all play important roles in the relative quality of the resulting metric.
For example, a metric trained on Twitter data may also perform well at evaluating dialogs generated for Reddit data, since the two datasets are constructed from online forums.
This section introduces the characteristics of the datasets used to train the various model-based metrics.

DailyDialog~\cite{li-etal-2017-dailydialog} is a human-written dialog dataset where the conversation topics are about day-to-day life. 

PersonaChat~\cite{zhang-etal-2018-personalizing} is a dataset that consists of persona-conditioned dialogs where each participant is assigned a persona and the goal is to become familiar with the other individual.
In contrast to DailyDialog, dialogs in PersonaChat have a clear objective and are more engaging.

The ConvAI dataset~\cite{dinan2019second} is based on PersonaChat with modifications to pre-processing and additional training examples.

TopicalChat~\cite{Gopalakrishnan2019} consists of knowledge-grounded human-human conversations, wherein two individuals have a conversation grounded on `interesting facts'. Models trained on TopicalChat are expected to be able to use external knowledge and have realistic knowledge-grounded conversations.

Twitter \cite{ritter-etal-2011-data} and Ubuntu Dialogue \cite{lowe-etal-2015-ubuntu} both result from a crawl of the internet.
Ubuntu Dialogue has very technical conversations regarding computer systems, while Twitter covers a broad, general set of topics. 

These datasets were generally used to train the metrics through the use of self-supervised objectives. 

\subsection{Referenced vs Reference-Free}

Word-overlap metrics are ineffective for dialog \citep{liu2016not} largely due to the one-to-many nature of dialog~\cite{zhao-etal-2017-learning}.
While this could be mitigated by using multiple reference responses, it is infeasible to collect a sufficiently large dataset to thoroughly cover the space of potential responses.
Thus, reference-free metrics have been proposed to circumvent the one-to-many problem.
Amongst the metrics assessed here, there are several reference-free evaluation metrics: HolisticEval, MAUDE, GRADE, USR, FED, FlowSore, USL-H, QuestEval, DEB, DynaEval, PredictiveEngage and DialogRPT.

In contrast to the referenced metrics, which compare the generated response to the reference response, reference-free metrics model the semantics of the dialog context and the generated response in order to reason about the response within the context of the dialog history. 

\subsection{Metrics not Assessed in this Paper}

Table~\ref{tab:not_used_metrics} lists recently proposed dialog metrics that are not assessed in this paper. There are several reasons a metric was not assessed: 
\begin{itemize}[noitemsep]
    \item The metric was not designed specifically for dialog. While most of these metrics were not included, a few metrics that fall into this category (e.g., BERTScore, QuestEval, and BLEURT) were assessed here, as a baseline and to represent this category.
    \item There was no released code, data, or pretrained model for reproducing their results. 
    \item The core idea of the metric is very similar to metrics that were assessed.
    \item The metric is infeasible to assess in our experimental setting, as it requires additional annotations. This may include requiring human annotations or information about the response generation models.
\end{itemize}

Some of the unassessed metrics that share ideas that are covered by other metrics assessed in this paper are: 

Embedding Methods \cite{kiros2015skip, liu-etal-2016-evaluate, rodriguez2021automatic} compute the similarity of system responses with the human reference through embeddings, which is equivalent to the approach used by RUBER and BERTScore. 

Language Model Evaluator \cite{nedelchev-etal-2020-language} evaluates dialog using the language model likelihood, which is identical to the approaches of USR and HolisticEval.

\section{Testing Datasets}

Table~\ref{tab:dataset_overview} shows statistics for the datasets used for testing. These datasets consist of human quality annotations. Concretely, each sample in the dataset consists of a dialog context, a generated response and a quality score. Optionally, a ground-truth response may also be included.

In general, these datasets are constructed using the following steps:
\begin{enumerate} [noitemsep]
    \item Choose an existing dialog dataset.
    \item Train a response generation model on the chosen dialog dataset.
    \item Generate responses for the validation/test set of the chosen dataset 
    \item Collect human quality annotations for the generated responses. 
\end{enumerate}
For data that does not contain referenced responses, only reference-free metrics were assessed.

The characteristics of the quality-annotated data significantly influence the performance of the metrics since dialog metrics might be originally developed and trained on data in a very different domain than the test domain.
Therefore, the following section briefly discusses different aspects of the testing data used here.

\subsection{Data Collection}

This section provides an overview of the different quality-annotated dialog datasets. Many of these datasets were collected in different settings. For example, as described in Section~\ref{sec:train_data}, DailyDialog consists of casual conversations about daily life while TopicalChat consists of knowledge-grounded conversations.
These differences influence various aspects of the data, such as the formality and complexity of the sentence structure. However, since it is not easy to quantify the complexity of the sentence structure, the average length and the number of distinct words in the dialog context (Ctx), human reference (Ref) and model hypothesis (Hyp) are used in Table~\ref{tab:dataset_overview}.
The length of the context and reference response in DailyDialog is shorter than on TopicalChat, which is likely influenced by the simpler topics in DailyDialog.

GRADE-DailyDialog has a significantly longer dialog history than the other two quality-annotated datasets that use DailyDialog. This is because those two datasets, HolisticEval and PredictiveEngage, only use one utterance of the dialog while GRADE uses two utterances.
On the other hand, responses in PredictiveEngage-DailyDialog use a larger number of distinct words, which is due to the fact that PredictiveEngage-DailyDialog also contains human-written responses in addition to responses generated by NLG systems.

While most of the dialog data was collected by recruiting human annotators via Amazon Mechanical Turks (AMT), the DSTC6 data \cite{DSTC6_End-to-End_Conversation_Modeling} uses dialogs from Twitter.
Therefore, we can expect the DSTC6 data to be noisier and more realistic. It poses a challenge for both NLG models and evaluation metrics.

In contrast to other quality-annotated datasets in which the dialog contexts are conversations between humans and only the responses are generated by the system, the DSTC9 and FED datasets provide human-system dialogs that were collected in an interactive setting. The DSTC9 data \cite{gunasekara2020overview} was collected on the DialPort platform through direct interaction between real users and open-domain chit-chat systems. The FED data~\cite{mehri-eskenazi-2020-unsupervised} contains both human-human and human-system conversations released by \citet{adiwardana2020towards}.

Despite using the same underlying dialog dataset (e.g., DailyDialog), the complexity and quality of the response may differ significantly across the quality-annotated datasets depending on the response generation model that was used. If responses in a quality-annotated dataset are from both simple sequence-to-sequence (Seq2Seq) models \cite{cho-etal-2014-learning} and state-of-the-art language models such as DialoGPT, this quality-annotated dataset can assess whether the metric can distinguish between high quality and low quality responses.

Distinguishing between responses produced by systems of very different quality is easier, since the low quality responses may not follow grammar rules and make simple mistakes.
On the other hand, if responses are only generated by state-of-the-art systems, the task becomes harder because the metric needs to rank responses as to whether they are appropriate in a dialog context.

Responses in USR-TopicalChat, USR-PersonaChat, HolisticEval-DailyDialog and the DSTC6 data are generated by a relatively simple model, such as an LSTM language model (LSTM LM), LSTM or Transformer Seq2Seq model, or Memory Network.

On the other hand, responses in the quality-annotated data labeled by GRADE come from the Transformer Seq2Seq model, DialoGPT, and retrieval model using Transformer or BERT (Transformer/BERT Ranker), which have relatively better empirical performance.
Furthermore, FED and DSTC9 data use state-of-the-art dialog systems to generate responses.
Specifically, FED data incorporates two systems, Meena \cite{adiwardana2020towards} and Mitsuku\footnote{\url{https://medium.com/pandorabots-blog/mitsuku-wins-loebner-prize-2018-3e8d98c5f2a7}}, and the DSTC9 data uses dialog systems including PLATO \cite{bao-etal-2020-plato}.
These high quality responses make the data more challenging.

\subsection{Quality Annotation} \label{sec:data_annotate}

After generating responses for each dataset, human annotators labeled the quality of each response. While most quality-annotated datasets provide annotations for the overall score of a given response, HolisticEval-DailyDialog only labels the context coherence of a response.\footnote{HolisticEval also released the data for evaluating response fluency. However since the data does not clearly disambiguate between the dialog context and the corresponding system response, it was not used in this paper}.
In addition to the overall score, USR and FED data provide annotations for different dialog qualities such as whether the response is coherent or interesting. These fine-grained annotations allow for a more comprehensive analysis of metrics.

\begin{table*}[]
\tiny
\centering
    \begin{tabular}{l|ccccl}
    \toprule
    Dataset & Num. Samples & Avg. Utts. & Avg. Ctx/Ref/Hyp Words & Distinct Ctx/Ref/Hyp Words & Used NLG models\\
    \midrule
    USR-TopicalChat & 300 & 12.2 & 236.3 / 25.3 / 22.4 & 2379 / 571 / 1018 & Transformers\\
    USR-PersonaChat & 240 & 9.3 & 98.4 / 11.7 / 12.1 & 1320 / 291 / 609 & Transformer Seq2Seq, LSTM LM, Memory Network\\
    GRADE-ConvAI2 & 600 & 2.0 & 23.4 / 12.1 / 11.3 & 1342 / 841 / 1100 & Transformer Seq2Seq, DialoGPT, BERT/Transformer Ranker  \\
    GRADE-DailyDialog & 300 & 2.0 & 24.5 / 12.9 / 10.8 & 1126 / 674 / 796 & Transformer Seq2Seq, Transformer Ranker\\
    HolisticEval-DailyDialog & 200 & 1.0 & 14.3 / - / 14.7 & 847 / - / 875 & LSTM Seq2Seq\\
    PredictiveEngage-DailyDialog & 600 & 1.0 & 12.0 / - / 13.7 & 1022 / - / 2475 & LSTM Seq2Seq\\
    GRADE-EmpatheticDialogue \shortcite{rashkin-etal-2019-towards} & 300 & 2.0 & 28.0 / 17.8 / 15.6 & 1430 / 1098 / 701 & Transformer Seq2Seq, Transformer Ranker \\
    DSTC6 \shortcite{DSTC6_End-to-End_Conversation_Modeling} & 40000 & 1.6 & 30.7 / 19.0 / 19.8  & 7399 / 3410 / 2473 & Systems based on LSTM Seq2Seq\\
    FED \shortcite{mehri-eskenazi-2020-unsupervised}  & 500 & 11.8 & 86.5 / - / 11.8 & 3628 / - / 1859 & Meena, Mitsuku\\
    DSTC9 \shortcite{gunasekara2020overview}  & 2200 & 27.2 & 231.7 / - / 10.6 & 35591 / - / 4839 & State-of-the-art systems including Plato and DialoGPT\\
    \end{tabular}
    \caption{Statistics of quality-annotated datasets.}
    \label{tab:dataset_overview}
\end{table*}

\begin{table*}[]
    \small
    \centering
    \begin{tabular}{lcccccccc}
    \toprule
        & \multicolumn{4}{c}{USR-TopicalChat} & \multicolumn{4}{c}{USR-PersonaChat}  \\
        & \multicolumn{2}{c}{Turn-Level} & \multicolumn{2}{c}{System-Level}  & \multicolumn{2}{c}{Turn-Level} & \multicolumn{2}{c}{System-Level}   \\
        & P & S  & P & S  & P & S   & P & S  \\
    \midrule
BLEU-4 & 0.216 & 0.296 & 0.874* & 0.900 & 0.135 & 0.090* & 0.841* & 0.800* \\
METEOR & 0.336 & 0.391 & 0.943 & 0.900 & 0.253 & 0.271 & 0.907* & 0.800* \\
ROUGE-L & 0.275 & 0.287 & 0.814* & 0.900 & 0.066* & 0.038* & 0.171* & 0.000* \\
ADEM & -0.060* & -0.061* & 0.202* & 0.700* & -0.141 & -0.085* & 0.523* & 0.400* \\
BERTScore & 0.298 & 0.325 & 0.854* & 0.900 & 0.152 & 0.122* & 0.241* & 0.000* \\
BLEURT & 0.216 & 0.261 & 0.630* & 0.900 & 0.065* & 0.054* & -0.125* & 0.000* \\
QuestEval & 0.300 & 0.338 & 0.943 & \textbf{1.000} & 0.176 & 0.236 & 0.885* & \textbf{1.000}\\
RUBER & 0.247 & 0.259 & 0.876* & \textbf{1.000} & 0.131 & 0.190 & \textbf{0.997} & \textbf{1.000} \\
BERT-RUBER & 0.342 & 0.348 & \textbf{0.992} & 0.900 & 0.266 & 0.248 & 0.958 & 0.200* \\
PONE & 0.271 & 0.274 & 0.893 & 0.500* & 0.373 & 0.375 & 0.979 & 0.800* \\
MAUDE & 0.044* & 0.083* & 0.317* & -0.200* & 0.345 & 0.298 & 0.440* & 0.400* \\
DEB & 0.180 & 0.116 & 0.818* & 0.400* & 0.291 & 0.373 & 0.989 & \textbf{1.000}\\
GRADE & 0.200 & 0.217 & 0.553* & 0.100* & 0.358 & 0.352 & 0.811* & \textbf{1.000} \\
DynaEval & -0.032* & -0.022* & -0.248* & 0.100* &  0.149 & 0.171 & 0.584* & 0.800*\\
USR & \textbf{0.412} & \textbf{0.423} & 0.967 & 0.900 & 0.440 & 0.418 & 0.864* & \textbf{1.000} \\
USL-H & 0.322 & 0.340 & 0.966 & 0.900 & \textbf{0.495} & \textbf{0.523} & 0.969 & 0.800*  \\
DialogRPT  & 0.120 & 0.105* & 0.944 & 0.600*& -0.064* & -0.083* & 0.347* & 0.800*\\
Deep AM-FM & 0.285 & 0.268 & 0.969 & 0.700* & 0.228 & 0.219 & 0.965 & \textbf{1.000} \\
HolisticEval & -0.147 & -0.123 & -0.919 & -0.200* & 0.087* & 0.113* & 0.051* & 0.000* \\
PredictiveEngage & 0.222 & 0.310 & 0.870* & 0.900 & -0.003* & 0.033* & 0.683* & 0.200* \\
FED & -0.124 & -0.135 & 0.730* & 0.100* & -0.028* & -0.000* & 0.005* & 0.400* \\
FlowScore & 0.095* & 0.082* & -0.150* & 0.400* & 0.118* & 0.079* & 0.678* & 0.800* \\
FBD & - & - & 0.916 & 0.100* & - & - & 0.644* & 0.800* \\
    \end{tabular}
    
\caption{Results on the USR data. All values are statistically significant to $p < 0.05$, unless marked by *.}
\label{tab:usr_data}
\end{table*}

\begin{table*}[]
    \tiny
    \centering
    \begin{tabular}{lcccccccccccc}
    \toprule
        & \multicolumn{4}{c}{GRADE-ConvAI2} & \multicolumn{4}{c}{GRADE-DailyDialog } & \multicolumn{4}{c}{GRADE-EmpatheticDialogue}  \\
        & \multicolumn{2}{c}{Turn-Level} & \multicolumn{2}{c}{System-Level}  & \multicolumn{2}{c}{Turn-Level} & \multicolumn{2}{c}{System-Level} & \multicolumn{2}{c}{Turn-Level} & \multicolumn{2}{c}{System-Level}  \\
        & P & S  & P & S  & P & S   & P & S  & P & S & P & S\\
    \midrule
BLEU-4 & 0.003* & 0.128 & 0.034* & 0.000* & 0.075* & 0.184 & \textbf{1.000*} & \textbf{1.000} & -0.051* & 0.002* & \textbf{1.000*} & \textbf{1.000} \\
METEOR & 0.145 & 0.181 & 0.781* & 0.600* & 0.096* & 0.010* & -1.000* & -1.000 & 0.118 & 0.055* & \textbf{1.000*} & \textbf{1.000} \\
ROUGE-L & 0.136 & 0.140 & 0.209* & 0.000* & 0.154 & 0.147 & \textbf{1.000*} & \textbf{1.000} & 0.029* & -0.013* & \textbf{1.000*} & \textbf{1.000} \\
ADEM & -0.060* & -0.057* & -0.368* & -0.200* & 0.064* & 0.071* & \textbf{1.000*} & \textbf{1.000} & -0.036* & -0.028* & \textbf{1.000*} & \textbf{1.000} \\
BERTScore & 0.225 & 0.224 & 0.918* & 0.800* & 0.129 & 0.100* & -1.000* & -1.000 & 0.046* & 0.033* & \textbf{1.000*} & \textbf{1.000} \\
BLEURT & 0.125 & 0.120 & -0.777* & -0.400* & 0.176 & 0.133 & \textbf{1.000*} & \textbf{1.000} & 0.087* & 0.051* & \textbf{1.000*} & \textbf{1.000} \\
QuestEval & 0.279 & 0.319 & 0.283* & 0.400* & 0.020* & 0.006* & -1.000* & -1.000 & 0.201 & 0.272 & \textbf{1.000*} & \textbf{1.000}  \\
RUBER & -0.027* & -0.042* & -0.458* & -0.400* & -0.084* & -0.094* & -1.000* & -1.000 & -0.078* & -0.039* & \textbf{1.000*} & \textbf{1.000} \\
BERT-RUBER & 0.309 & 0.314 & 0.885* & \textbf{1.000} & 0.134 & 0.128 & -1.000* & -1.000 & 0.163 & 0.148 & \textbf{1.000*} & \textbf{1.000} \\
PONE & 0.362 & 0.373 & 0.816* & 0.800* & 0.163 & 0.163 & -1.000* & -1.000 & 0.177 & 0.161 & \textbf{1.000*} & \textbf{1.000} \\
MAUDE & 0.351 & 0.304 & 0.748* & 0.800* & -0.036* & -0.073* & \textbf{1.000*} & \textbf{1.000} & 0.007* & -0.057* & \textbf{1.000*} & \textbf{1.000} \\
DEB & 0.426 & 0.504 & \textbf{0.995} & \textbf{1.000} & \textbf{0.337} & \textbf{0.363} & 1.000* & 1.000 & \textbf{0.356} & \textbf{0.395} & \textbf{1.000*} & \textbf{1.000}\\
GRADE & \textbf{0.566} & \textbf{0.571} & 0.883* & 0.800* & 0.278 & 0.253 & -1.000* & -1.000 & 0.330 & 0.297 & \textbf{1.000*} & \textbf{1.000} \\
DynaEval & 0.138 & 0.131 &  -0.996 & -1.000 & 0.108* & 0.120  & -1.000* & -1.000 & 0.146 & 0.141 & -1.000* & -1.000\\
USR & 0.501 & 0.500 & \textbf{0.995} & \textbf{1.000} & 0.057* & 0.057* & -1.000* & -1.000 & 0.264 & 0.255 & \textbf{1.000*} & \textbf{1.000} \\
USL-H & 0.443 & 0.457 & 0.971 & 1.000 & 0.108* & 0.093* & -1.000* & -1.000 & 0.293 & 0.235 & \textbf{1.000*} & \textbf{1.000} \\
DialogRPT & 0.137 & 0.158 &  -0.311* & -0.600* & -0.000* & 0.037* & -1.000* & -1.000 & 0.211 & 0.203 & 1.000* & 1.000 \\
Deep AM-FM & 0.117 & 0.130 & 0.774* & 0.400* & 0.026* & 0.022* & \textbf{1.000*} & \textbf{1.000} & 0.083* & 0.058* & \textbf{1.000*} & \textbf{1.000} \\
HolisticEval & -0.030* & -0.010* & -0.297* & -0.400* & 0.025* & 0.020* & \textbf{1.000*} & \textbf{1.000} & 0.199 & 0.204 & -1.000* & -1.000 \\
PredictiveEngage & 0.154 & 0.164 & 0.601* & 0.600* & -0.133 & -0.135 & -1.000* & -1.000 & -0.032* & -0.078* & \textbf{1.000*} & \textbf{1.000} \\
FED & -0.090 & -0.072* & -0.254* & 0.000* & 0.080* & 0.064* & \textbf{1.000*} & \textbf{1.000} & -0.014* & -0.044* & \textbf{1.000*} & \textbf{1.000} \\
FlowScore & - & - & - & - & - & - & - & - & - & - & - & - \\
FBD & - & - & -0.235* & -0.400* & - & - & -1.000* & -1.000 & - & - & -1.000* & -1.000 \\
    \end{tabular}
    
\caption{Results on the GRADE data. All values are statistically significant to $p < 0.05$, unless marked by *.}
\label{tab:grade_data}
\end{table*}

\begin{table*}[]
    \small
    \centering
    \begin{tabular}{lcccc}
    \toprule
    & \multicolumn{4}{c}{DSTC6} \\
    & \multicolumn{2}{c}{Turn-Level} & \multicolumn{2}{c}{System-Level} \\ 
    & P & S & P & S \\
    \midrule
BLEU-4 & 0.131 & 0.298 & -0.064* & 0.050* \\
METEOR & 0.307 & 0.323 & 0.633 & 0.084* \\
ROUGE-L & 0.332 & 0.326 & 0.487 & 0.215* \\
ADEM & 0.151 & 0.118 & 0.042* & 0.347* \\
BERTScore & \textbf{0.369} & \textbf{0.337} & 0.671 & 0.265* \\
BLEURT & 0.326 & 0.294 & 0.213* & 0.426* \\
QuestEval & 0.188 & 0.242  & -0.215* & 0.206*  \\
RUBER & 0.114 & 0.092 & -0.074* & 0.104* \\
BERT-RUBER & 0.204 & 0.217 & \textbf{0.825} & 0.093* \\
PONE & 0.208 & 0.200 & 0.608 & 0.235* \\
MAUDE & 0.195 & 0.128 & 0.739 & 0.217* \\
DEB & 0.211 & 0.214 & -0.261* & 0.492 \\
GRADE & 0.119 & 0.122 & 0.784 & 0.611 \\
DynaEval & 0.286 & 0.246 & 0.342* & -0.050* \\
USR & 0.184 & 0.166 & 0.432* & 0.147* \\
USL-H & 0.217 & 0.179 & 0.811 & 0.298* \\
DialogRPT & 0.170 & 0.155 & 0.567 & 0.334* \\
Deep AM-FM & 0.326 & 0.295 & 0.817 & \textbf{0.674} \\
HolisticEval & 0.001* & -0.004* & 0.010 & -0.002 \\
PredictiveEngage & 0.043 & 0.004* & -0.094* & -0.409* \\
FED & -0.106 & -0.083 & 0.221* & 0.322* \\
FlowScore & 0.064 & 0.095 & 0.352* & 0.362* \\
FBD & - & - & -0.481 & -0.234* \\
    \end{tabular}
\caption{Results on the DSTC6 data. All values are statistically significant to $p < 0.05$, unless marked by *.}
\label{tab:dstc6}
\end{table*}

\begin{table*}[]
    \small
    \centering
    \begin{tabular}{lcccccccccc}
    \toprule
    & \multicolumn{2}{c}{PE-DailyDialog} &  \multicolumn{4}{c}{FED}  & \multicolumn{4}{c}{DSTC9} \\
    &  \multicolumn{2}{c}{Turn-Level}  &   \multicolumn{2}{c}{Turn-Level} & \multicolumn{2}{c}{Dialog-Level}  & \multicolumn{2}{c}{Dialog-Level} & \multicolumn{2}{c}{System-Level} \\
    & P & S & P & S & P & S & P & S & P & S \\
    \midrule
    
    QuestEval & 0.296 & 0.341 & 0.037* & 0.093* & -0.032* & 0.080* & 0.026* & 0.043  & 0.604 & 0.527* \\
    MAUDE & 0.104 & 0.060* & 0.018* & -0.094* & -0.047* & -0.280 & 0.059 & 0.042* & 0.224* & 0.045* \\
    DEB & 0.516 & 0.580 & 0.230 & 0.187 & -0.130* & 0.006* & 0.085 & 0.131 & 0.683 & 0.473* \\
    GRADE & 0.600 & 0.622 & 0.134 & 0.118 & -0.034* & -0.065* & -0.078 & -0.070 & -0.674 & -0.482* \\
    DynaEval & 0.167 & 0.160 & \textbf{0.319} & \textbf{0.323} & \textbf{0.503} & \textbf{0.547} & 0.093 & 0.101 & 0.652 & 0.727 \\
    USR & 0.582 & 0.640 & 0.114 & 0.117 & 0.093* & 0.062* & 0.019* & 0.020* & 0.149* & 0.127* \\
    USL-H & \textbf{0.688} & \textbf{0.699} & 0.201 & 0.189 & 0.073* & 0.152* & 0.105 & 0.105 & 0.566* & 0.755  \\
    DialogRPT &  0.489 & 0.533 & -0.118 & -0.086* & -0.221 & -0.214 & 0.076 & 0.069 &  0.685 & 0.555* \\
HolisticEval & 0.368 & 0.365 & 0.122 & 0.125 & -0.276 & -0.304 & 0.015* & 0.002* & -0.019* & -0.100* \\
PredictiveEngage & 0.429 & 0.414 & 0.024* & 0.094* & 0.026* & 0.155* & 0.114 & 0.115 & 0.809 & 0.664 \\
FED & 0.164 & 0.159 & 0.120 & 0.095 & 0.222 & 0.320 & 0.128 & 0.120 & 0.559* & 0.391* \\
FlowScore & - & - & -0.065* & -0.055* & -0.073* & -0.003* & \textbf{0.147} & \textbf{0.140} & \textbf{0.907} & \textbf{0.900} \\
FBD & - & - & - & - & - & -  & - & - & -0.669 & -0.627 \\
    \end{tabular}
    
\caption{Results on PredictiveEngage-DailyDialog, FED, and DSTC9 data. Here, we denote PredictiveEngage-DailyDialog by PE-DailyDialog due to space constraint. All values are statistically significant to $p < 0.05$, unless marked by *.}
\label{tab:engage_fed_dstc9}
\end{table*}

\begin{table}[]
    \small
    \centering
    \begin{tabular}{lcc}
    \toprule
    & \multicolumn{2}{c}{HolisticEval-DailyDialog} \\
    & P & S  \\
    \midrule
    QuestEval & 0.285 & 0.260 \\
    MAUDE & 0.275 & 0.364  \\
    DEB & 0.584 & 0.663 \\
    GRADE & \textbf{0.678} & 0.697  \\
    DynaEval & -0.023* & -0.009* \\
    USR & 0.589 & 0.645  \\
    USL-H & 0.486 & 0.537 \\
    DialogRPT & 0.283 & 0.332 \\
HolisticEval & 0.670 & \textbf{0.764} \\
PredictiveEngage & -0.033* & 0.060* \\
FED & 0.485 & 0.507  \\
FlowScore & - & - \\
FBD & - & - \\
    \end{tabular}

\caption{Results on HolisticEval-DailyDialog. All values are statistically significant to $p < 0.05$, unless marked by *.}
\label{tab:holistic}
\end{table}

\begin{table*}[]
    \small
    \centering
    \begin{tabular}{lcccc}
    \toprule
     & \multicolumn{2}{c}{Generative} & \multicolumn{2}{c}{Retrieval} \\
    & P & S & P & S \\
    \midrule
BLEU-4 & 0.100 & 0.128 & 0.023* & 0.236 \\
METEOR & 0.130 & 0.086 & 0.129 & 0.185 \\
ROUGE-L & 0.134 & 0.108 & 0.212 & 0.238 \\
ADEM & -0.117 & -0.118 & -0.045* & -0.070* \\
BERTScore & 0.182 & 0.185 & 0.175 & 0.186 \\
BLEURT & 0.153 & 0.147 & 0.108 & 0.093 \\
QuestEval & 0.115 & 0.112 & 0.105 & 0.190 \\
RUBER & 0.022* & 0.031* & -0.092 & -0.109 \\
BERT-RUBER & 0.158 & 0.159 & 0.206 & 0.208 \\
PONE & 0.227 & 0.236 & 0.191 & 0.192 \\
MAUDE & 0.126 & 0.161 & 0.072* & 0.031* \\
DEB & 0.281 & 0.313 & 0.307 & 0.420 \\
GRADE & \textbf{0.300} & \textbf{0.314} & \textbf{0.435} & \textbf{0.428} \\
DynaEval & 0.110 & 0.113 & 0.143 & 0.157 \\
USR & 0.130 & 0.149 & 0.402 & 0.415 \\
USL-H & 0.126 & 0.161 & 0.307 & 0.330 \\
DialogRPT & -0.082 & -0.067 & -0.009* & 0.001* \\
Deep AM-FM & 0.068 & 0.068 & 0.086 & 0.085 \\
HolisticEval & -0.037* & -0.003* & 0.070* & 0.066* \\
PredictiveEngage & -0.063* & -0.060* & 0.010* & 0.028* \\
FED & -0.072 & -0.095 & -0.169 & -0.184 \\
FlowScore & 0.236 & 0.215 & -0.059* & -0.053* \\
FBD & - & - & - & - \\
    \end{tabular}
    
\caption{Results on evaluating different response generation models. All values are statistically significant to $p < 0.05$, unless marked by *.}
\label{tab:model_type}
\end{table*}

\begin{figure*}
\begin{subfigure}{.5\textwidth}
    \centering
    \includegraphics[width=\linewidth]{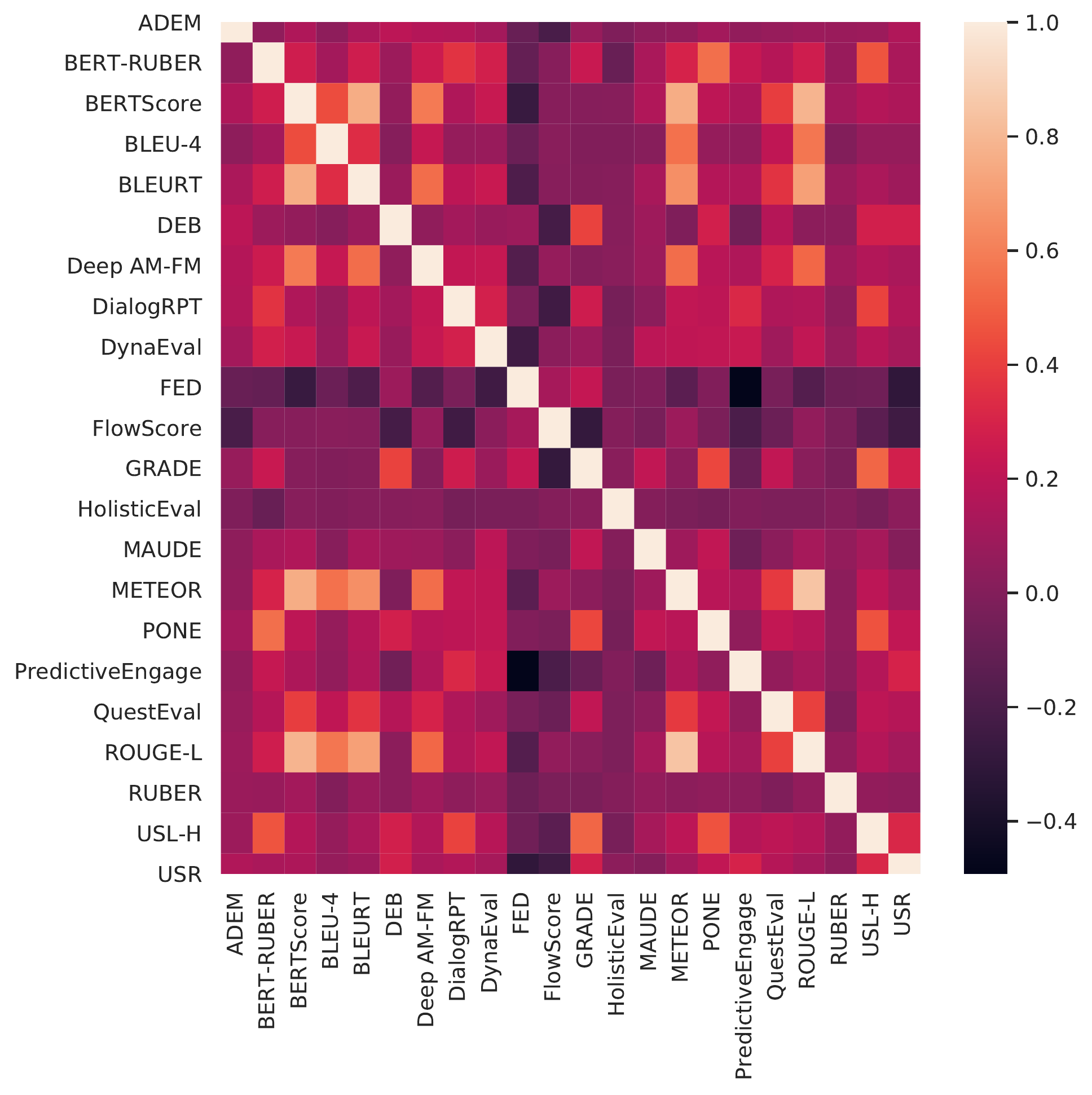}
    \caption{Pearson}
\end{subfigure}
\begin{subfigure}{.5\textwidth}
    \centering
    \includegraphics[width=\linewidth]{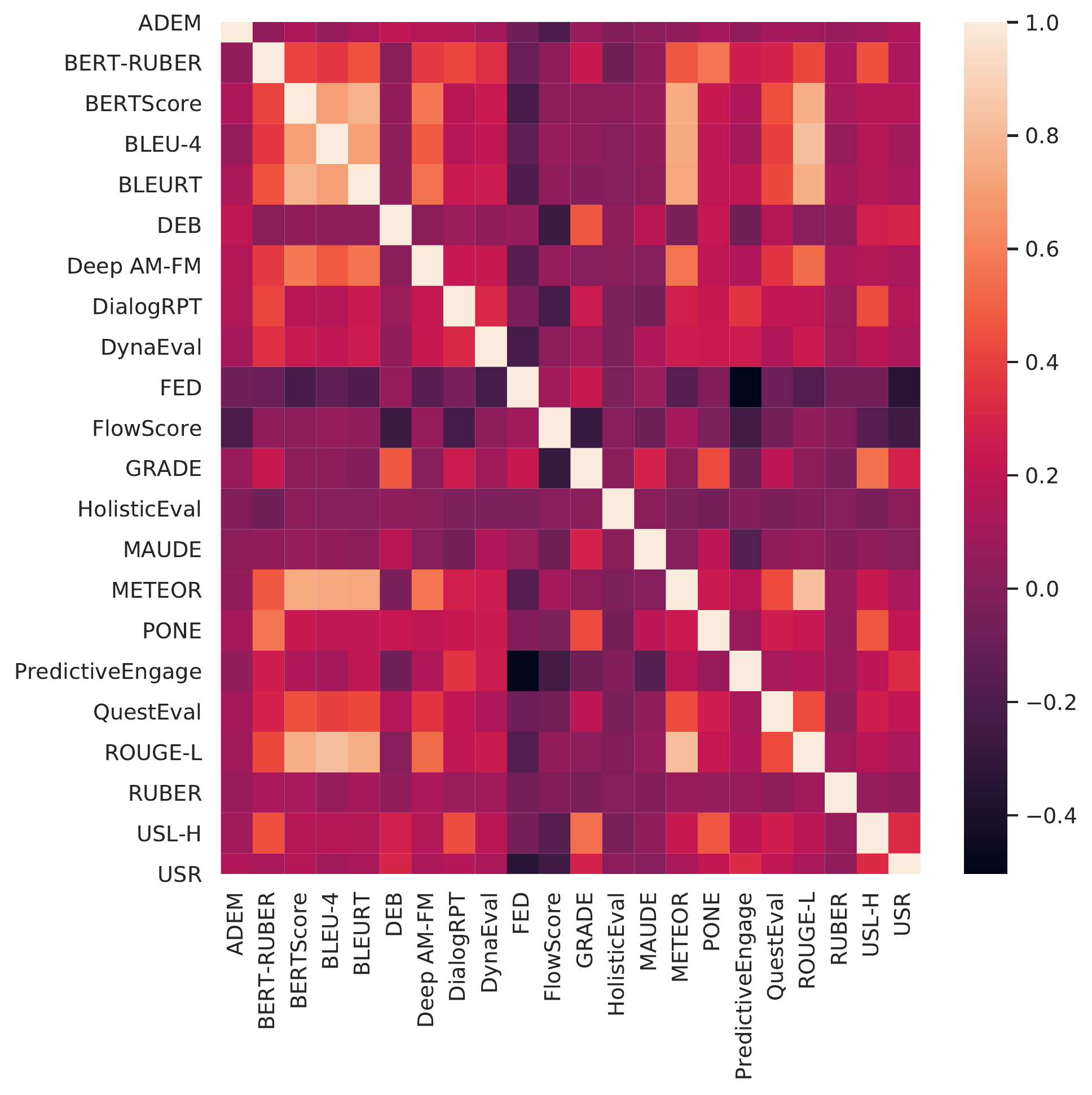}
    \caption{Spearman}
\end{subfigure}
\caption{Pearson and Spearman correlation between different system outputs. We use datasets with human references to compute scores.}
\label{fig:metric_correlation}
\end{figure*}

\begin{figure*}
\begin{subfigure}{.5\textwidth}
    \centering
    \includegraphics[width=\linewidth]{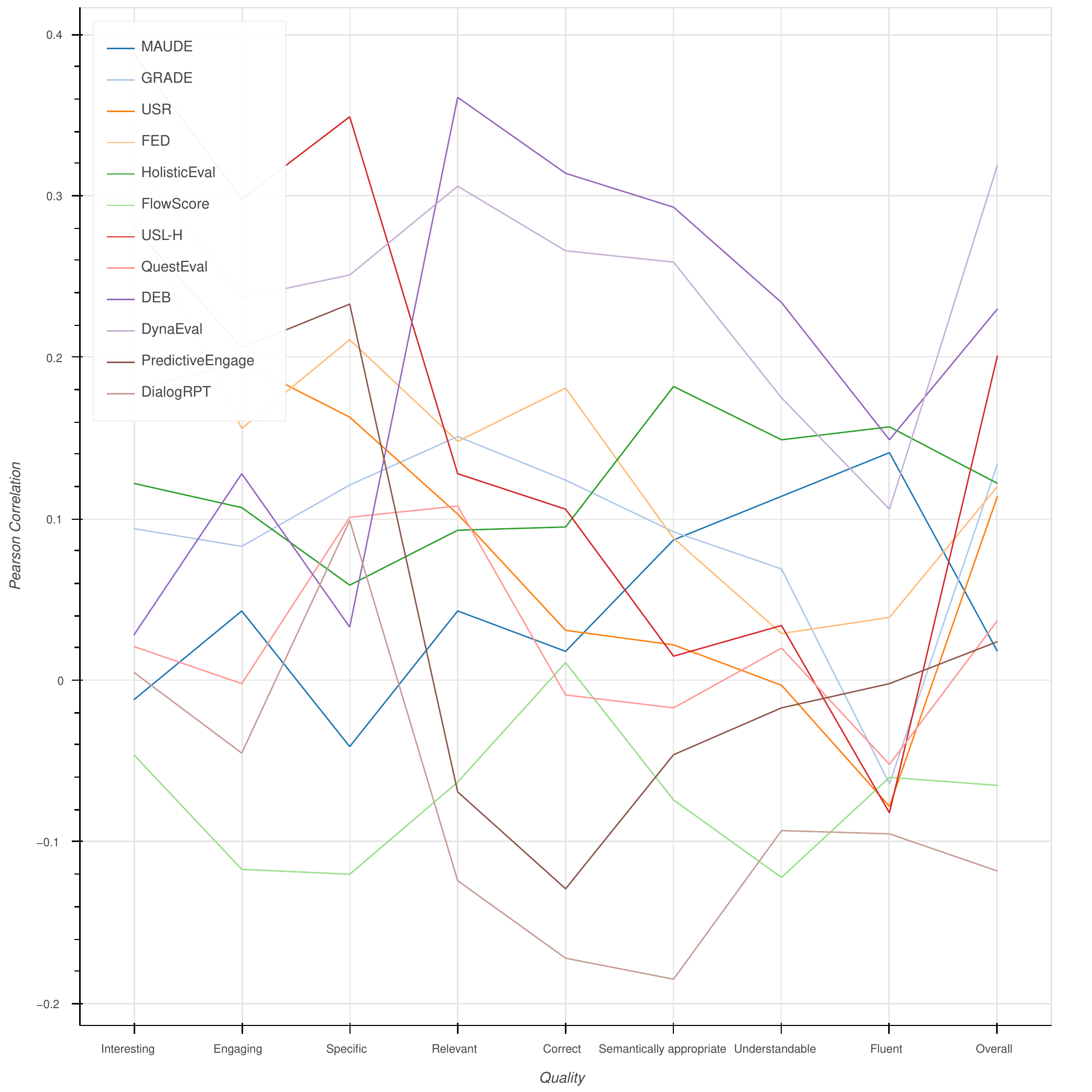}
    \caption{Pearson}
\end{subfigure}
\begin{subfigure}{.5\textwidth}
    \centering
    \includegraphics[width=\linewidth]{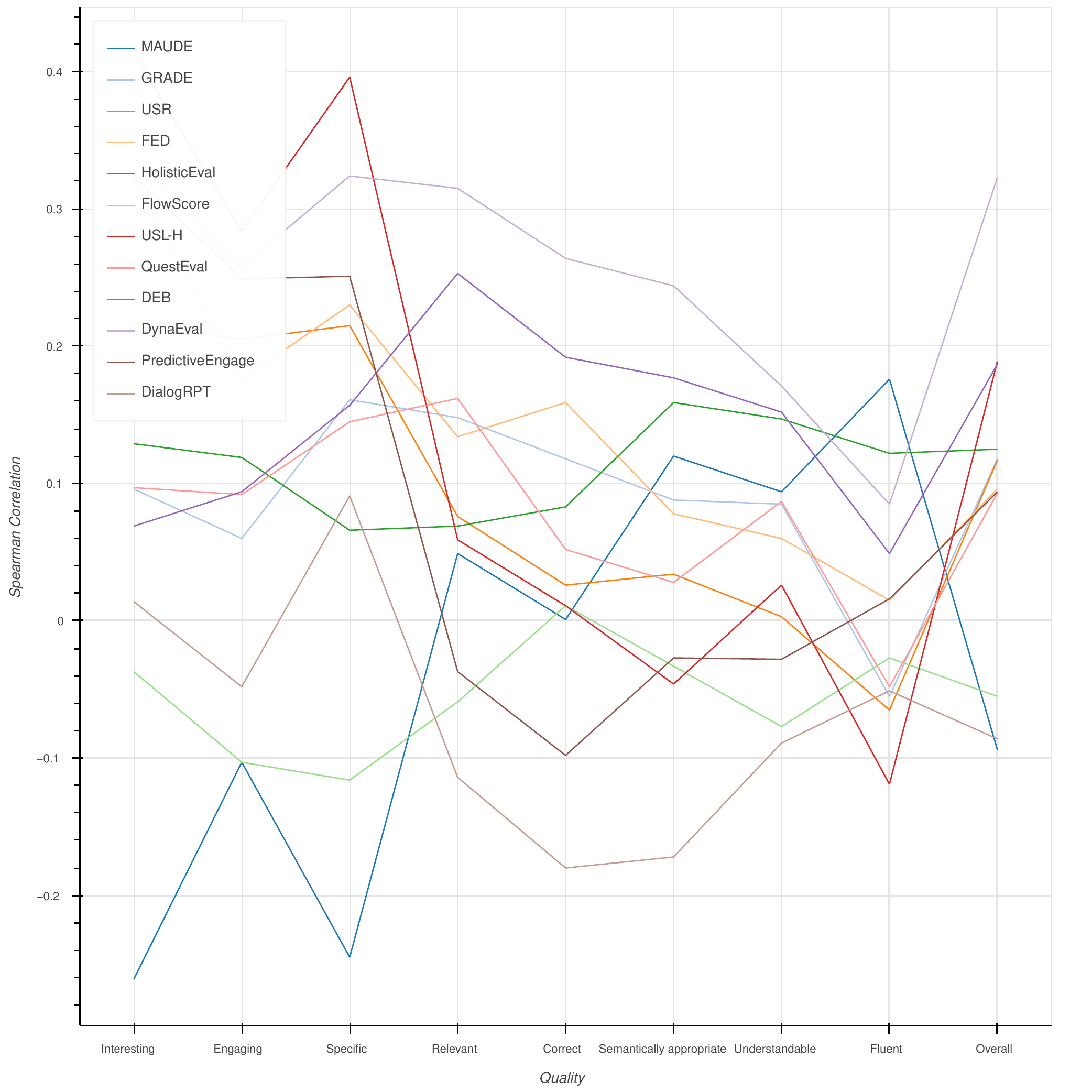}
    \caption{Spearman}
\end{subfigure}
\caption{Pearson and Spearman correlation to different turn-level annotation qualities on the FED data.}
\label{fig:turn_aspects}
\end{figure*}

\begin{figure*}
\begin{subfigure}{.5\textwidth}
    \centering
    \includegraphics[width=\linewidth]{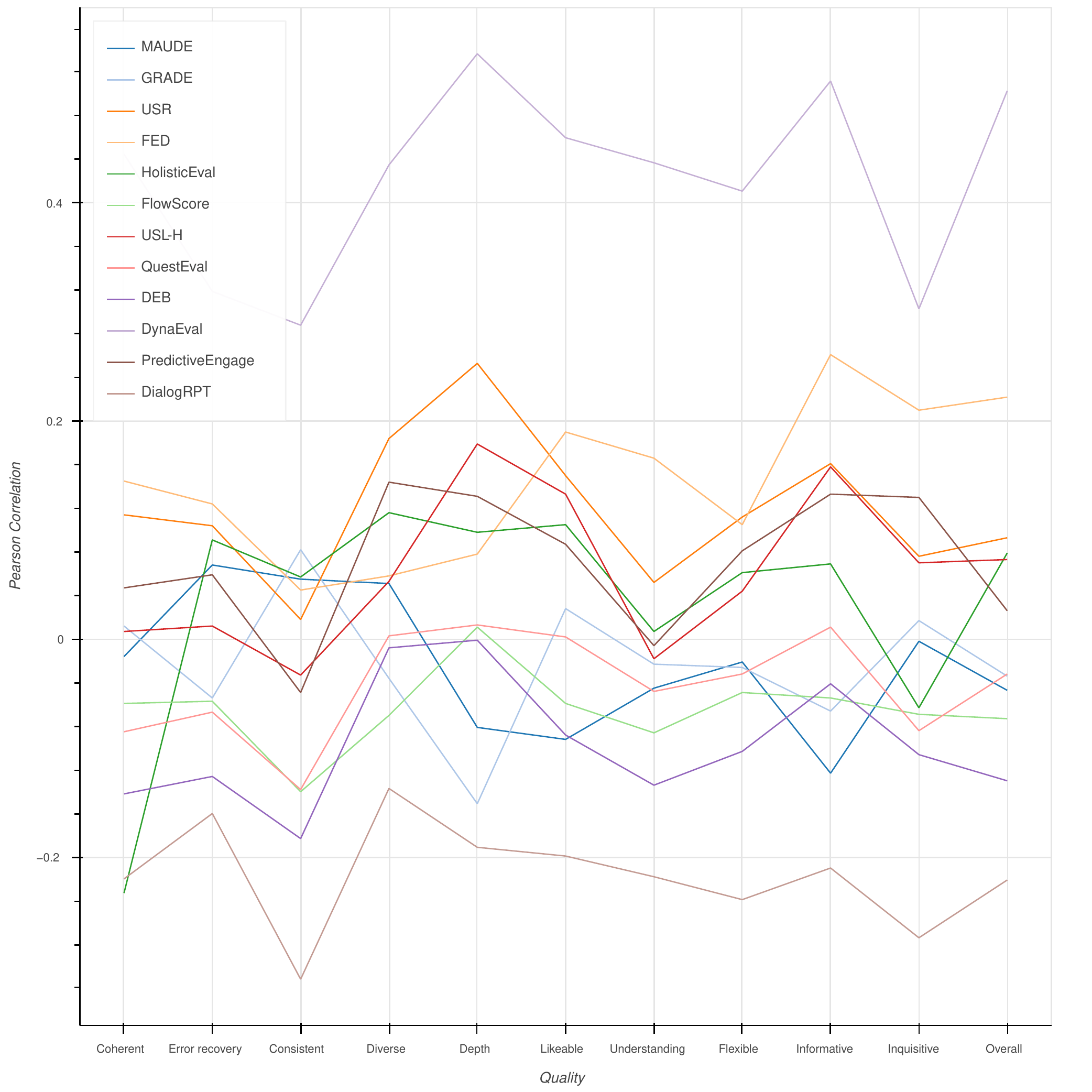}
    \caption{Pearson}
\end{subfigure}
\begin{subfigure}{.5\textwidth}
    \centering
    \includegraphics[width=\linewidth]{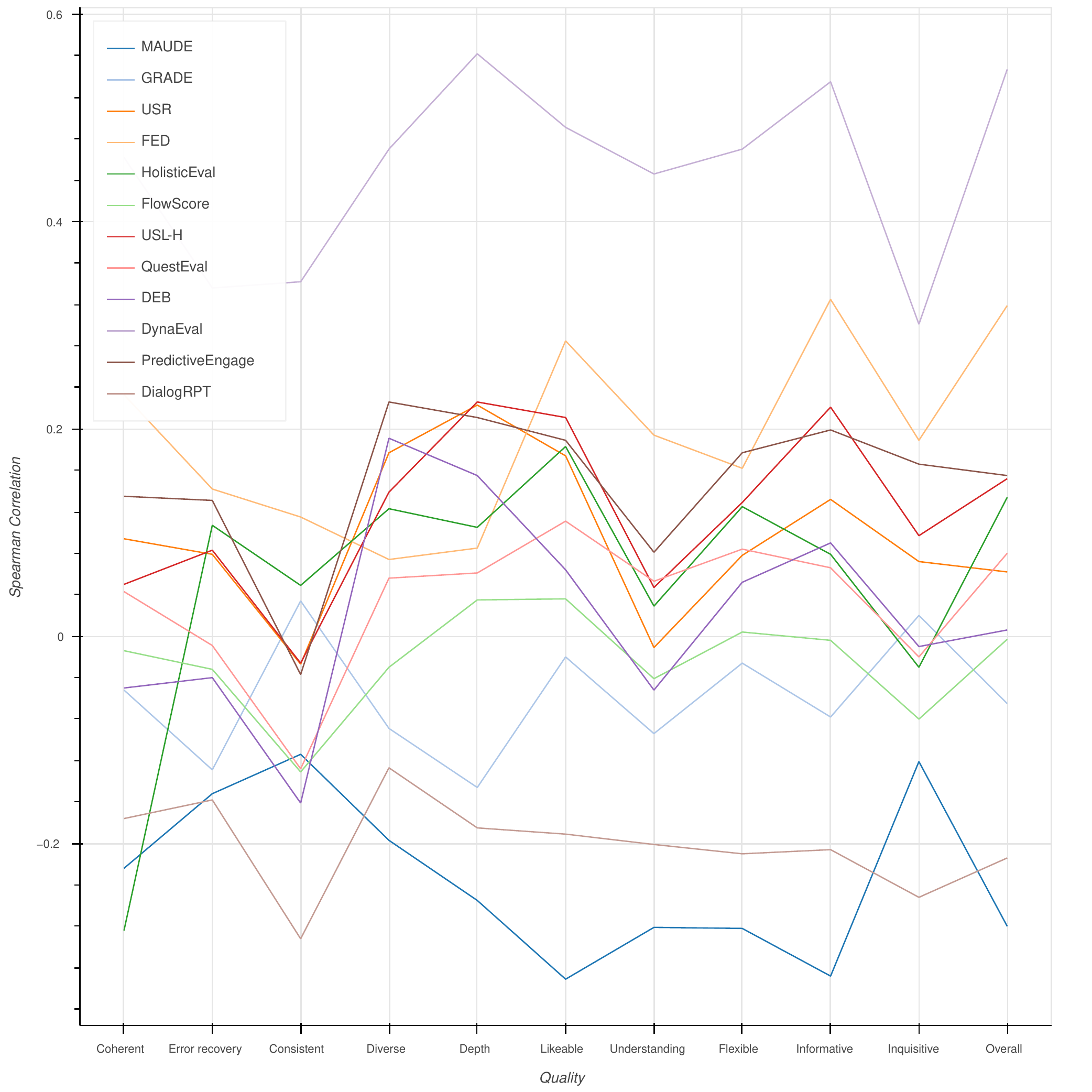}
    \caption{Spearman}
\end{subfigure}
\caption{Pearson and Spearman correlation to different dialog-level annotation qualities on the FED data.}
\label{fig:dialog_aspects}
\end{figure*}

\begin{figure*}
\begin{subfigure}{.5\textwidth}
    \centering
    \includegraphics[width=\linewidth]{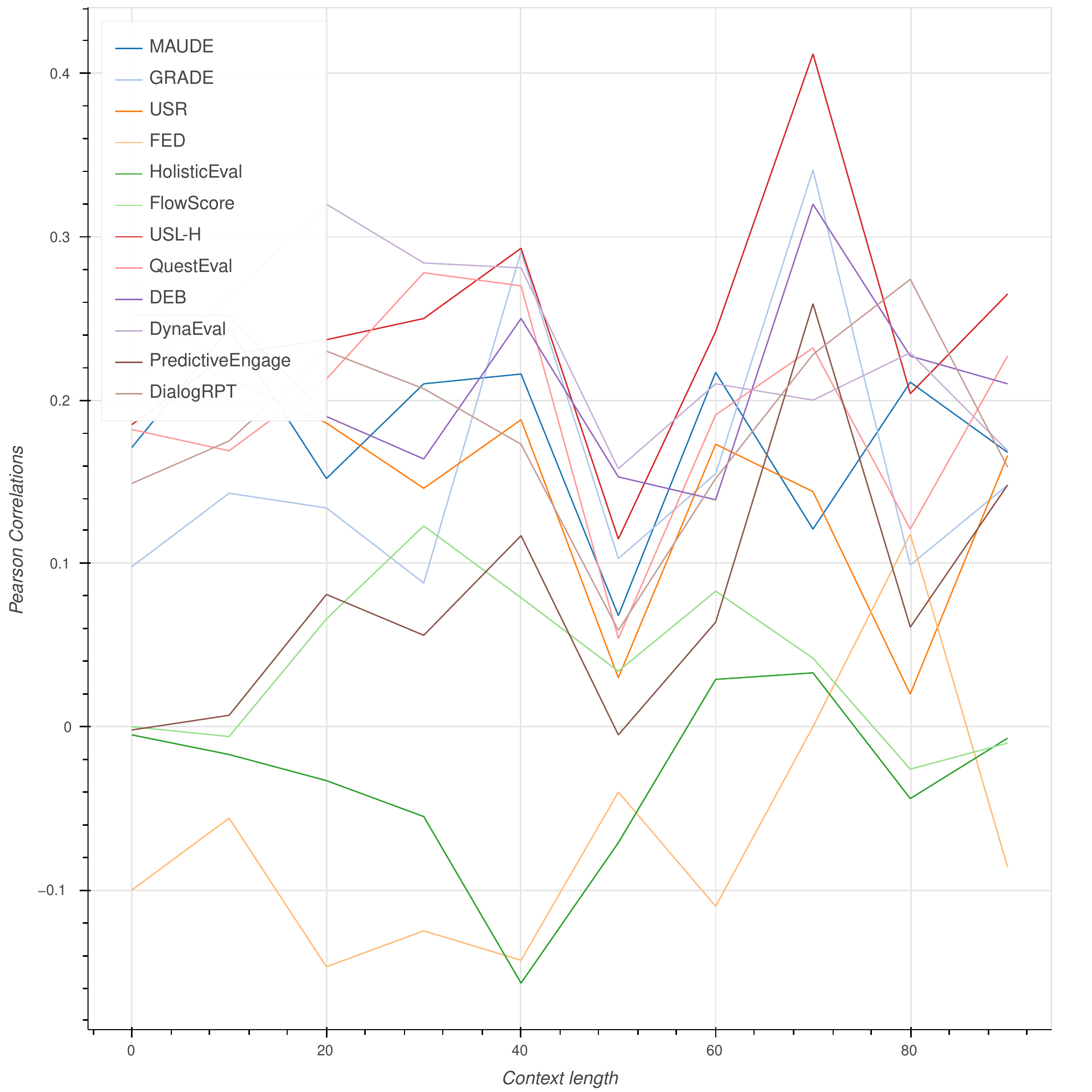}
    \caption{Pearson}
\end{subfigure}
\begin{subfigure}{.5\textwidth}
    \centering
    \includegraphics[width=\linewidth]{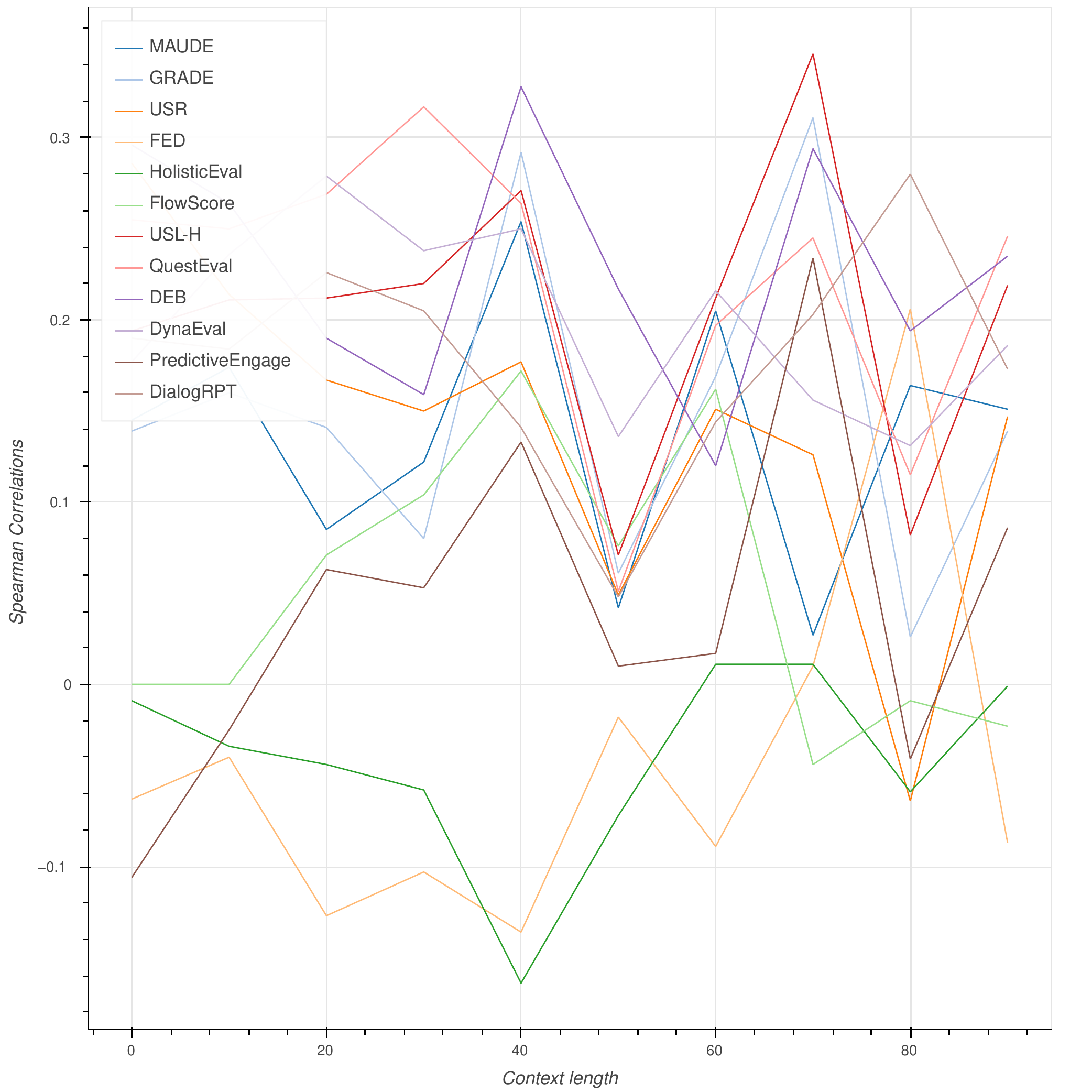}
    \caption{Spearman}
\end{subfigure}
\caption{Pearson and Spearman correlation when varying the dialog context length. }
\label{fig:ref_free_ctx_length}
\end{figure*}

\begin{table*}[]
    \tiny
    \centering
    \begin{tabular}{lcccccccccccc}
    \toprule
        & \multicolumn{2}{c}{USR-TopicalChat} & \multicolumn{2}{c}{USR-PersonaChat} & \multicolumn{2}{c}{GRADE-ConvAI2} & \multicolumn{2}{c}{GRADE-DailyDialog} & \multicolumn{2}{c}{GRADE-EmpatheticDialogue} & 
        \multicolumn{2}{c}{DSTC6} \\
         & P & S  & P & S  & P & S  & P & S  & P & S  & P & S \\
         \midrule

PredictiveEngage & 0.222 & 0.310 & -0.003* & 0.033* & 0.154 & 0.164 & -0.133 & -0.135 & -0.032* & -0.078* & 0.043 & 0.004* \\
+ RUBER & 0.283 & 0.327 & 0.057* & 0.096* & 0.105 & 0.102 & -0.149 & -0.156 & -0.065* & -0.055* & 0.103 & 0.055 \\
+ PONE & 0.308 & 0.350 & 0.218 & 0.209 & 0.333 & 0.339 & -0.034* & -0.035* & 0.077* & 0.069* & 0.156 & 0.122 \\
USR & 0.412 & 0.423 & 0.440 & 0.418 & 0.501 & 0.500 & 0.057* & 0.057* & 0.264 & 0.255 & 0.184 & 0.166 \\
+ GRADE & 0.424 & 0.432 & 0.456 & 0.430 & 0.523 & 0.528 & 0.081* & 0.078* & 0.281 & 0.267 & 0.191 & 0.173 \\
+ USL-H & 0.429 & 0.440 & 0.468 & 0.451 & 0.517 & 0.522 & 0.062* & 0.062* & 0.278 & 0.265 & 0.197 & 0.178 \\
+ DEB & 0.359 & 0.362 & 0.453 & 0.473 & 0.552 & 0.551 & 0.222 & 0.190 & 0.391 & 0.385 & 0.248 & 0.234 \\
+ GRADE + PONE & 0.435 & 0.447 & 0.473 & 0.452 & 0.535 & 0.539 & 0.089 & 0.086 & 0.289 & 0.270 & 0.206 & 0.187 \\
+ GRADE + PONE + PredictiveEngage & 0.447 & 0.468 & 0.468 & 0.449 & 0.534 & 0.538 & 0.072* & 0.070* & 0.275 & 0.261 & 0.202 & 0.180 \\
+ GRADE + USL-H + DEB & 0.377 & 0.382 & 0.476 & 0.495 & \textbf{0.571} & \textbf{0.579} & 0.229 & 0.198 & \textbf{0.399} & 0.387 & 0.254 & 0.238 \\
GRADE & 0.200 & 0.217 & 0.358 & 0.352 & 0.566 & 0.571 & 0.278 & 0.253 & 0.330 & 0.297 & 0.119 & 0.122 \\
+ PONE & 0.282 & 0.297 & 0.435 & 0.436 & 0.547 & 0.556 & 0.275 & 0.279 & 0.306 & 0.279 & 0.191 & 0.179 \\
+ PONE + PredictiveEngage & 0.335 & 0.348 & 0.373 & 0.366 & 0.515 & 0.530 & 0.118 & 0.117 & 0.219 & 0.186 & 0.181 & 0.155 \\
+ USL-H & 0.302 & 0.304 & 0.493 & 0.480 & 0.551 & 0.556 & 0.229 & 0.237 & 0.342 & 0.307 & 0.185 & 0.169 \\
+ DEB & 0.201 & 0.217 & 0.319 & 0.403 & 0.463 & 0.549 & \textbf{0.345} & \textbf{0.325} & 0.368 & 0.391 & 0.214 & 0.200 \\
+ DEB + USL-H & 0.219 & 0.270 & 0.347 & 0.465 & 0.484 & 0.547 & 0.339 & 0.311 & 0.377 & \textbf{0.395} & 0.225 & 0.215 \\
USL-H & 0.322 & 0.340 & 0.495 & 0.523 & 0.443 & 0.457 & 0.108* & 0.093* & 0.293 & 0.235 & 0.217 & 0.179 \\
+ DEB & 0.201 & 0.269 & 0.322 & 0.482 & 0.452 & 0.501 & 0.333 & 0.258 & 0.366 & 0.390 & 0.224 & 0.223 \\
All-Metrics & \textbf{0.459} & \textbf{0.473} & \textbf{0.522} & \textbf{0.534} & 0.566 & 0.561 & 0.163 & 0.149 & 0.366 & 0.349 & \textbf{0.288} & \textbf{0.268} \\
    \end{tabular}
\caption{Results of different combinations of metrics. All values are statistically significant to $p < 0.05$, unless marked by *. The last row (All-Metrics), is the average of all the metrics.}
\label{tab:combined}
\end{table*}

\section{Experiments}

This section describes the assessment of the evaluation metrics. Wherever possible, the released pretrained model was used to reproduce results. Since RUBER, BERT-RUBER and PONE do not release their pretrained models, those models were trained on DailyDialog. For USR, we use the released model which is trained on TopicalChat. For PredictiveEngage, the original paper combined the model with RUBER for the best performance. In this assessment only the engagement score model is used for comparison to the other metrics. The potential combination of it with various other metrics is explored in Section~\ref{sec:combined_metrics}.

FlowScore only scores dialogs with more than 3 utterances, so it cannot be used on some of the quality-annotated data. 
For FBD, we use a different data preprocessing pipeline on the USR and GRADE data to ensure a fair comparison, and as such the results may differ from those reported in the FBD paper.
Note that HolisticEval was designed to generate fine-grained scores for four aspects:  \textit{context coherence}, \textit{language fluency}, \textit{response diversity}, and \textit{logical self-consistency}.
The \textit{response diversity} metric requires access to the response generation model used to produce a response. However, since access to the original response generation models for each of the quality-annotated datasets is not available, the \textit{response diversity} metric cannot be calculated. Thus for HolisticEval, the average of the other three metrics was used to compute the overall score for a response.

\subsection{Results on Datasets with Human Reference}

This section describes the performance of the metrics on the quality-annotated datasets that contain a human reference. This distinction is important, because the referenced metrics (e.g., BLEU, METEOR, BERTScore) can only be assessed on these datasets. Table~\ref{tab:usr_data}, Table~\ref{tab:grade_data} and Table~\ref{tab:dstc6} present the correlations of the metrics on the USR, GRADE and DSTC6 data, respectively.

Rule-based metrics perform surprisingly well on USR-TopicalChat,  USR-PersonaChat and DSTC6. However, they fall short on the GRADE data. This may be due to the fact that the responses in the GRADE data are produced by better NLG models, as described in Section~\ref{sec:data_annotate}. 
Since rule-based metrics calculate word-overlap metrics between the system responses and a human reference, they can consistently detect poor responses which only contain irrelevant words.
However, the rule-based metrics struggle to score responses from state-of-the-art dialog systems since those responses require a deeper semantic understanding of the dialog context and the generated response.

By incorporating large pretrained models, BERTScore and BLEURT have similar performance to the rule-based metrics when assessed on the USR and DSTC6 data, and they perform better on the GRADE data.
The use of pretrained models allows BERTScore and BLEURT to better capture the semantic content of the generated responses.
As such, while BERTScore calculates recall and precision in the same manner as traditional rule-based metrics, the use of a pretrained model allows it to handle more challenging responses, like those present in the GRADE data.

A similar improvement is observed when comparing RUBER and BERT-RUBER.
BERT-RUBER uses a pretrained BERT model and achieves a significantly higher correlation over the RUBER metric, which only uses pretrained Word2Vec vectors \cite{mikolov2013efficient}. This further stresses the importance of using contextualized word representations in evaluation metrics.

PONE improves upon BERT-RUBER by incorporating data augmentation and weighted negative sampling techniques. The improvement shown on the GRADE data proves that PONE is better able to model semantic content and therefore performs better at scoring the higher-quality responses in the GRADE data.

MAUDE improves upon the unreferenced metric in BERT-RUBER by training with contrastive loss. Compared to BERT-RUBER, MAUDE performs better on USR-PersonaChat and GRADE-ConvAI2 and has worse performance on the other quality-annotated datasets. This is probably due to the fact that MAUDE was trained on PersonaChat and therefore performs better on quality-annotated data from the same domain.

The performance of DEB demonstrates the importance of training data in model-based metrics.
DEB replaces the BERT pretraining data with a dialog corpus, and further finetunes the model on a manually-created dataset consisting of adversarial responses designed for the NSP objective.
DEB achieves the highest performance on GRADE-DailyDialog and GRADE-EmpatheticDialogue, which indicates the effectiveness of the constructed dataset.

The USR and GRADE metrics have high performance on their respective datasets. This phenomenon probably occurs for several reasons. First, both metrics fine-tune pretrained models in a self-supervised manner on certain dialog datasets. It is expected that the metrics would perform better on quality-annotated datasets corresponding to the data they were fine-tuned on. For example, the USR metric which was fine-tuned on TopicalChat should perform better on USR-TopicalChat. Second, the metrics may be designed for the specific attributes of the data. For example, the choice of the topic graph of GRADE may have been influenced by phenomena observed in the GRADE data. Finally, metrics may be optimized specifically for the data (i.e., through hyperparameter tuning on a validation set). This observation stresses the importance of testing on a variety of different quality-annotated datasets in order to ensure the generality of an evaluation metric and to avoid over-optimizing to specific datasets.

USR performs well on the GRADE data, particularly on GRADE-ConvAI2, which is similar to USR-PersonaChat. USR aggregates the results produced by three different self-supervised models that approximate different qualities of dialog (fluency, relevance and knowledge grounding). Since the dialog retrieval model is, in principle, similar to the unreferenced metric used in RUBER, this result demonstrates that using a language model results in an improvement when evaluating high-quality responses.
On the other hand, GRADE does not show very strong performance on the USR or DSTC6 data. Since rule-based metrics already perform very well on these quality-annotated datasets, it may be the case that the topic transition flow in GRADE does not help when assessing lower-quality responses.

USL-H has a similar performance trend to USR, which is reasonable since the two metrics both combine response selection models and language models.
USL-H outperforms USR on the USR-PersonaChat dataset.
In contrast to USR, USL-H incorporates the VUP model which validates whether a given response is grammatically correct.
These results demonstrate the effectiveness of the VUP model in identifying poor responses.

Deep AM-FM was trained on data crawled from Twitter. It performs relatively well on the DSTC6 data which was also obtained from Twitter. Similar to the aforementioned observations regarding USR and GRADE performance, Deep AM-FM results stress the importance of training a metric on the target evaluation domain. However, Deep AM-FM does not outperform BERTScore on the DSTC6 data. This is likely due to the way DSTC6 was collected. The response generation models used to generate responses in the DSTC6 data are relatively simple, relative to the models used in the USR and GRADE datasets. Therefore, as described above, word-overlap metrics (especially sophisticated word-overlap metrics like BERTScore) will be especially strong at evaluating this type of data. Deep AM-FM achieves low correlations on the GRADE data, likely due to the domain mismatch.
Since the RNN language model used in Deep AM-FM is trained with Twitter data, it might not be able to deal with the responses written by human annotators.

FED, HolisticEval and PredictiveEngage are designed for evaluating specific, fine-grained qualities of dialog. Simply taking the average of the fine-grained scores produced by these metrics obtains poor correlation with the overall quality annotations. However, we can not simply conclude that these metrics are poorly designed. The metrics do indeed have strong correlation to human scores on specific qualities, which will be discussed below.

\subsection{Results on Datasets without Human Reference}

This section describes the results for the datasets that do not contain a reference response. Only reference-free metrics can be evaluated on these quality-annotated datasets.

Table~\ref{tab:engage_fed_dstc9} presents results of the reference-free metrics on HolisticEval-DailyDialog, FED, and DSTC9 data.

In the FED dataset, there are two types of annotations: turn-level and dialog-level. Turn-level annotations assess the quality of a single response, while dialog-level annotations assess the entire dialog.
Similar to the GRADE data, the responses in the FED data are generated by state-of-the-art dialog systems \citep{adiwardana2020towards}, which makes the data particularly challenging.

On turn-level annotations in the FED data, the best performing metrics are DynaEval, USL-H, and DEB. In particular, DynaEval performs considerably better than other metrics. The strong performance of DynaEval is a consequence of the fact that DynaEval considers the dependencies between pairs of utterances, by using a graph-based representation of the dialog. Through this, DynaEval is able to measure how well-formed a dialog is and therefore assess the quality of a given system utterance.

Many metrics which perform well on the USR and GRADE data, do not do as well on the FED data. This may be because of the longer dialog context in the FED data. While the average number of words in the context of GRADE-ConvAI2 is 23.4, the context of FED has on average 86.5 words and 11.8 utterances. If a model has not seen long contexts at training time, it will struggle to perform well on longer contexts at test time.

On the dialog-level annotations in the FED data, FED and DynaEval have the best performance. FED and DynaEval are both specifically designed to evaluate the dialog quality while other reference-free metrics only assess turn-level quality. DynaEval achieves a much stronger correlation than all the other metrics. While FED directly uses the pretrained DialoGPT model without any supervision, DynaEval is trained to leverage a learned graph representation to assess the quality of a dialog. Other than DynaEval and FED, PredictiveEngage has a higher dialog-level correlation than the other metrics. Since PredictiveEngage explicitly predicts how engaging a response is, the higher performance suggests that "engaging" is an important factor for dialog-level evaluation. HolisticEval performs significantly worse when evaluating dialog-level correlation than it does on turn-level correlation. This degradation of performance may be due to the fact that HolisticEval only fine-tunes the GPT-2 model with language modeling objectives and they may be insufficient when the model needs to take the whole dialog into consideration.

Similar to the FED data, the DSTC9 data also contains interactive human-system dialogs from state-of-the-art dialog systems. Again, the longer dialog contexts in combination with the use of state-of-the-art response generation models makes DSTC9 a particularly challenging dataset for evaluation. PredictiveEngage, DynaEval, FED, USL-H and FlowScore have the highest dialog-level correlations on the DSTC9 dataset. 

DynaEval, FED and PredictiveEngage achieve strong dialog-level correlations on both the FED and the DSTC9 data. This may occur for two reasons. First, these models are designed to capture properties that are better aligned with dialog-level annotations: DynaEval measures coherence using a graph-based representation of the dialog, PredictiveEngage measures engagement, FED measures a number of different dialog-level qualities (e.g., coherence, topic depth, etc.). Second, the model architectures of PredictiveEngage and FED are relatively simple. This may make the two metrics less sensitive to longer dialog contexts. Though DynaEval, PredictiveEngage and FED do well at assessing state-of-the-art dialog systems in the FED and DSTC9 data, they underperform on the GRADE and USR data. This suggests that these two metrics are optimal for dialog-level evaluation, but less so for turn-level evaluation.

FlowScore obtains the strongest performance on the DSTC9 data, however, it underperforms on other quality-annotated datasets including the FED data. This suggests that FlowScore is particularly good at modelling long and complex dialogs.

\subsection{Results on System-Level Correlation}

This section presents the system-level correlation on the different quality-annotated datasets. System-level correlation is a strong indication of the ability of a metric to rank several response generation models. The different datasets cover a varying number of systems: USR-TopicalChat (5 systems), USR-PersonaChat (4 systems), GRADE-ConvAI2 (4 systems), GRADE-DailyDialog (2 systems), GRADE-EmpatheticDialog (2 systems), DSTC6 (11 systems) and DSTC9 (20 systems). GRADE-DailyDialog and GRADE-EmpatheticDialog only consist of 2 systems, which makes system-level correlation much less informative.

In general, metrics that have a higher turn-level correlation tend to have a higher system-level correlation. Interestingly, some metrics which perform poorly on turn-level correlation do much better on system-level correlation, suggesting that averaging out over all the examples produced by a dialog system reduces the noise in the metric's scores. For example, GRADE has a comparable system-level correlation to the best-performing Deep AM-FM on the DSTC6 data, despite having a lower turn-level correlation.

On the other hand, BERTScore, which has the highest turn-level correlation on the DSTC6 data, performs poorly on system-level correlation.
This suggests that word-overlap metrics may struggle to accurately assess entire systems.

In general, many metrics attain strong system-level correlation suggesting that they can reasonably distinguish between systems of different qualities. The DSTC6 and DSTC9 data contain the largest number of systems, making it difficult to attain high system-level correlation. Deep AM-FM attains high system-level correlation on the DSTC6 data and FlowScore attains high system-level correlation on the DSTC9 data. 

\subsection{Performance on Various Dialog Qualities}

Since dialog quality is inherently multi-faceted \cite{walker-etal-1997-paradise, see-etal-2019-makes}, it is inadequate to only evaluate dialog metrics on the overall response score.
Therefore, this paper also presents the correlation of the metrics with various dialog qualities in Table~\ref{tab:holistic}, Figure~\ref{fig:turn_aspects}, and Figure~\ref{fig:dialog_aspects}. 

As described in Section~\ref{sec:data_annotate}, HolisticEval-DailyDialog annotates the context coherence of responses. In the FED data, there are 8 fine-grained turn-level qualities and 10 fine-grained dialog-level qualities.
For metrics that produce fine-grained scores, we use the corresponding fine-grained score to measure correlation. For example, because PredictiveEngage evaluates the engaging quality of the dialog, the \textit{engaging} score, rather than the overall score, is used to measure correlation with the \textit{engaging} quality in the FED dataset.

Different dialog metrics perform well on different qualities. On the HolisticEval-DailyDialog data, the best performing metric is HolisticEval. USL-H performs the best on measuring the \textit{specific} quality in the FED data.

On the other hand, HolisticEval performs poorly at measuring coherence on the FED data. One possible reason could be that the language modeling objective of HolisticEval makes it insufficient for measuring dialog-level qualities on the FED data. Another explanation is that the HolisticEval-DailyDialog data and the FED data may interpret \textit{context coherent} differently .

Similarly, GRADE and USR are highly correlated measuring coherence on HolisticEval-DailyDialog data, but they have relatively poor performance on the coherent quality of dialog-level annotation in the FED data.
This further suggests that the definitions of \textit{context coherent} in HolisticEval-DailyDialog and \textit{coherent} in the FED data may differ. HolisticEval-DailyDialog do not provide their precise annotation instructions, and as such the definitions of coherence between the two datasets cannot be compared.

USR and USL-H both do well predicting context coherence and engaging on the the HolisticEval-DailyDialog and FED data, respectively. These two metrics also have good performance on the USR and GRADE data. In contrast, HolisticEval and PredictiveEngage do well on measuring coherence and engaging, but underperform on the USR and GRADE data. This suggests that while fine-grained metrics are important qualities to measure, modelling \textit{only} these qualities is insufficient. Instead, it is better to design a metric that measures many different qualities and can aggregate them to form an overall score, such as the combination of MLM/dialog retrieval in USR and MLM/VUP/NSP in USL-H.

\subsection{Metric Output Similarities} \label{sec:output_sim}

To analyze the similarities between the metrics, we plot the correlation between metric outputs in Figure~\ref{fig:metric_correlation}.
As expected, there is a strong correlation between the outputs of BERTSCore, and BLEURT, the rule-based metrics. Interestingly, Deep AM-FM also is strongly correlated with these word-overlap metrics. This is surprising since Deep AM-FM was specifically designed for evaluating dialog while the others are intended for general-purpose NLG evaluation. This may be because the Deep AM-FM metric compares the generated response to the reference response and, as such, will favor generated responses that have high word-overlap with the reference response.

BERT-RUBER, PONE, GRADE, and USL-H are another group of metrics with similar behavior. Since BERT-RUBER, PONE, and GRADE share a common triplet rank loss and train on DailyDialog, it is not surprising that these three metrics behave similarly. Although USL-H is trained with different objectives, the same behavior might be due to the use of the same training data (DailyDialog) and the same pretrained model (BERT).
Moreover, USR has a slightly higher correlation to this group.
This is likely because USR aggregates multiple qualities similar to USL-H and models the relevance of responses, similar to the RUBER-based metrics and GRADE. 

\subsection{Effect of Context Length} \label{sec:context_len}

Most reference-free metrics compute their scores by comparing the dialog context with the response. Therefore, it is interesting to determine if these metrics perform differently according to the length of the dialog context. The quality-annotated samples are grouped by their context lengths with a length interval of 10. Figure~\ref{fig:ref_free_ctx_length} shows performance differences of the reference-free metrics at different context lengths.

With the exception of USL-H, DEB, HolisticEval, FED, PredictiveEngage, and DialogRPT, the metrics' performance decreases as the context lengthens. This may be due to the fact that those metrics are trained on shorter dialogs and struggle to understand longer dialog contexts. On the other hand, the performance of HolisticEval, DialogRPT, and FED increases as the context lengthens. These metrics incorporate GPT-2-based language models to score the responses, while the other metrics mostly rely on the BERT-based models. BERT-based models are optimized for local coherence on the BookCorpus and Wikipedia through the MLM objective and limit context length. Thus GPT-2-based metrics perform better on longer dialogs.

The performance of USL-H, DEB, and PredictiveEngage does not change much at different context lengths.
For PredictiveEngage, this could be because PredictiveEngage uses several MLP layers during the pooling of the pretrained BERT embeddings to predict the output. 
USL-H uses the VUP model which aims to determine if a response is gramatically correct. The use of the VUP model may make USL-H robust to longer context lengths. 
The BERT model used in DEB is pretrained on the Reddit corpus which may improve its ability to model long-range correlations in the dialog.
These different factors make metrics more robust to different context lengths.

\subsection{Performance on Different Response Generation Models}

This section presents the correlations of the metrics on both generative and retrieval response generation models (see Table~\ref{tab:model_type}). Most of the metrics perform similarly on these two types of models. However, USR, GRADE, and FlowScore perform very differently. USR and GRADE are especially good at scoring responses from the retrieval model. USR performs significantly less well at evaluating generative models while GRADE still achieves the highest correlation. In contrast, FlowScore performs well at evaluating generative models but performs poorly on generative models. One reason for this may be that responses from generative models are longer and more complex than the ones from retrieval models.

\subsection{Combining Metrics} \label{sec:combined_metrics}

Many metrics rely on an ensemble of different models. Inspired by this, this section explores the possibility that a combination of metrics may be successful (Table~\ref{tab:combined}). Since it is not feasible to exhaustively explore all possible combinations of the metrics, some combinations of the best-performing metrics were explored here. The metrics are combined through simple averaging. \textit{Future work should explore more sophisticated mechanisms for combining metrics.}

First, PredictiveEngage, RUBER and PONE were combined, as proposed in \cite{ghazarian2020predictive}. Indeed, the resulting combined metric shows significant improvement, especially on the USR-TopicalChat and USR-Personachat data, but correlation decreases on other datasets. 

Next, combinations of USR, GRADE, USL-H and DEB (the best performing metrics on the referenced quality-annotated datasets) were explored. The combination of USR with GRADE, USL-H, PONE and PredictiveEngage performs better on the USR data as well as on GRADE-ConvAI2. 
Given the strong performance of GRADE on the GRADE data, combining it with the other metrics negatively impacts its performance. However, combining GRADE and DEB does result in an improvement.
Interestingly, combining DEB with USR and USL-H also negatively affects the results.
In Figure~\ref{fig:metric_correlation}, we observe that the outputs of GRADE and DEB have high correlation to each other, which is an indication of the similar behavior of the two metrics.
The output similarities of metrics may be a good indicator for how to best combine metrics.

The last row of Table~\ref{tab:combined}, shows the result of taking the average of the scores of all of the metrics that were assessed in this paper.
While the idea of an \textit{All-Metric} metric is simple, and metrics are combined through simple averaging, the results are surprisingly good across the different referenced quality-annotated datasets.
This result highlights the potential of combining various evaluation metrics in some smart manner.

\section{Conclusions}

This paper provides a comprehensive assessment of various automatic evaluation metrics for dialog. Many different evaluation metrics have been proposed in recent years (2019 - 2021). To our knowledge there has not yet been a consistent comparison of these metrics. The goal of this paper is to make that comparison over a variety of different datasets and in a variety of settings.

The following are the high-level takeaways from the assessment and analysis of these metrics:

\begin{itemize}
    \item USR, GRADE, DEB, and USL-H are the best performing metrics for evaluating response generation (i.e., turn-level correlation). Their performance is consistently higher on the majority of different datasets.
    
    \item FlowScore, PredictiveEngage, DynaEval and FED perform the best on dialog-level evaluation on the FED and DSTC9 datasets.
    
    \item System-level correlations are generally very strong across a variety of metrics. Most impressively are the high correlations of Deep AM-FM on the DSTC6 data (11 different systems) and of FlowScore on the DSTC9 data (20 different systems). 
    
    \item Unsurprisingly, the majority of the metrics perform best on the datasets they were originally evaluated on. This suggests that the training data is important. Future work should explore strategies for transferring these metrics to new datasets (i.e., by fine-tuning them on in-domain data). This also suggests that the architectural design and choice of hyperparameters is influenced by the target dataset (either through direct optimization or human decision making). And so this further stresses the importance of assessing evaluation metrics on a variety of different quality-annotated datasets.
    
    \item Metrics also perform differently depending on the type of dialog system used to produce a response. Metric performance differs significantly between quality-annotated datasets with simple models (e.g., DSTC6, USR) and datasets with more sophisticated models (e.g., GRADE, FED, DSTC9). This suggests that there is not necessarily a one-size-fits-all metric, and the best metric depends on the type of data it is to deal with.
    
    \item Many metrics struggle with longer contexts, especially those that rely on BERT-based models. In contrast, metrics that use GPT-2-based models perform better. 
    
    \item Evaluating metrics on specific fine-grained qualities sheds much light on their underlying behavior. 
    
    \item In general, metrics combining different models have more consistently better performance. In a similar vein, combining different metrics yields strong results. A promising future direction is to develop more sophisticated strategies for combining evaluation metrics.
\end{itemize}

The results indicate several important directions for future work on the assessment of dialog evaluation metrics:

\begin{itemize}
    \item Many evaluation metrics rely on pretrained language models. The analysis in this paper, highlights the impacts of the choice of pretrained language model (e.g., GPT-2 better handles longer contexts while BERT-based models are better at measuring local coherence). Future work should further explore the impact of the choice of pretrained language models for evaluation.
    
    \item Many metrics are trained on specific dialog data (e.g., TopicalChat, DailyDialog, etc.). Unsurprisingly, metrics generally perform better on the data they were trained on. Since many metrics rely on self-supervised training, it is possible to further fine-tune them on other dialog datasets. Future work should explore adapting metrics to new domains through self-supervised fine-tuning on in-domain dialog data (e.g., fine-tune MAUDE on TopicalChat to attain better performance).
    
    \item Simple combinations of evaluation metrics yield promising results. Future work should explore more sophisticated mechanisms for combining different metrics. This may, for example, necessitate a mechanism (either a trained model or heuristic) for determining which metric is best suited to evaluate a particular response.
\end{itemize}

To facilitate the assessment of future automatic evaluation metrics for dialog, the code used for the assessment in this paper has been open-sourced for efficient assessment of metrics on a variety of different quality-annotated datasets: \url{https://github.com/exe1023/DialEvalMetrics}.

\section{Acknowledgements}

We thank the authors of the various metrics for releasing their code. This work is funded by National Science Foundation grant CNS-1512973. The opinions expressed in this paper do not necessarily reflect those of the National Science Foundation. 

\clearpage
\bibliography{anthology,custom}
\bibliographystyle{acl_natbib}

\end{document}